%% file: main.tex
\documentclass{article}

\usepackage[preprint]{neurips_2026}

\usepackage[utf8]{inputenc}
\usepackage[T1]{fontenc}
\usepackage{hyperref}
\usepackage{url}
\usepackage{booktabs}
\usepackage{multirow}
\usepackage{graphicx}
\usepackage{subcaption}
\usepackage{amsmath}
\usepackage{amssymb}
\usepackage{amsfonts}
\usepackage{amsthm}
\usepackage{nicefrac}
\usepackage{microtype}
\usepackage{xcolor}

\graphicspath{{figures/}}

\input{macros}

\title{Market-Conditioned JEPA: Learning from Unpredictable Market Shocks}

\author{
    Yuqi Wang$^{1}$ \quad Fengyuan Liu$^{1,2}$ \quad Mukai Li$^{1}$ \quad Haochen Luo$^{1}$ \quad Lei Li$^{1}$ \quad Qi Liu$^{1}$ \\
    $^{1}$The University of Hong Kong \quad $^{2}$Grace Investment Machine \\
    \texttt{wangyuqi@connect.hku.hk}, \texttt{liuqi@hku.hk}
}

\begin{document}

\maketitle

\begin{abstract}
Much of the variation in financial time series is driven by market-wide shocks that cannot be predicted from an asset's own history. Per-asset self-supervised learning treats this variation as irreducible noise. Under a latent factor model, we show that conditioning prediction on the realized market factor makes the asset's market exposure necessary for optimal prediction, while reducing the noise against which the asset-specific component is learned. Building on this insight, we introduce Market-Conditioned JEPA (MC-JEPA), a joint-embedding predictive architecture for financial time series. During training, we construct a leave-one-out market condition for each asset from the future target embeddings of other assets and incorporate it through factor loadings derived from the asset's own representation. On 11 years of minute-level Chinese A-share data, MC-JEPA improves over per-asset JEPA across nine targets spanning returns, market exposure, risk, and liquidity, with the strongest and most statistically significant gains in exposure, risk, and liquidity. It achieves the best RankIC on all liquidity targets and on return horizons from two hours to five days, outperforming time-series foundation models and self-supervised baselines, and delivers the best portfolio performance in a historical backtest. Ablations show that the return and exposure gains require the condition to be aligned with the realized market movement. After restoring the volatility scale removed by normalization, the predictor's factor loadings rank stocks by market beta nearly as well as a supervised probe, without labels. A synthetic factor-model study with known parameters shows that conditioning improves the recovery of both market exposure and the asset-specific component.
\end{abstract}

\input{sections/01_introduction}
\input{sections/02_related_work}
\input{sections/03_method}
\input{sections/04_experiments}
\input{sections/06_conclusion}

\input{sections/07_limitations}

\bibliographystyle{plainnat}
\bibliography{references}

\appendix
\input{sections/A_appendix}
\input{sections/B_appendix}
\input{sections/C_appendix}
\input{sections/D_appendix}
\input{sections/E_appendix}
\input{sections/F_appendix}


\end{document}

%% file: macros.tex
\newcommand{\R}{\mathbb{R}}
\newcommand{\E}{\mathbb{E}}
\newcommand{\sg}{\operatorname{sg}}

\newtheorem{proposition}{Proposition}

%% file: sections/01_introduction.tex
\section{Introduction}
\label{sec:intro}

Portfolio construction and risk management require estimates of an asset's return, volatility, liquidity, and market exposure, which measures how an asset responds to market-wide movements. Self-supervised learning offers a way to learn a unified representation that supports these diverse tasks. Recent work has developed a large number of time-series encoders and foundation models~\citep{Yuqietal-2023-PatchTST, yue2022ts2vec, pmlr-v235-goswami24a, pmlr-v235-das24c, ansari2025chronos2univariateuniversalforecasting}, including models pretrained specifically on financial data~\citep{shi2025kronosfoundationmodellanguage}. However, these methods largely learn from individual time series, overlooking the common-factor structure of financial markets.{\renewcommand{\thefootnote}{}\footnotetext{Code is available at \url{https://github.com/OpenFinMind/MC-JEPA}.}}

Under this structure, much of an asset's variation is driven by market-wide shocks~\citep{sharpe1964capital, fama1993common}, which are difficult to predict from an asset's own history. Therefore, a per-asset prediction objective treats the unpredictable component of these shocks as irreducible noise. This has two costs. First, the optimal predictor need only recover the asset's conditional mean, without identifying how the asset responds to market movements. This provides limited supervision for market exposure. Second, the asset-specific signal that the predictor learns is observed against this market noise. Existing approaches that explicitly model cross-asset relationships~\citep{xu2021hist, li2024master, gu2021autoencoder} are largely designed for supervised return prediction and often require cross-sectional information at inference. FactorVAE~\citep{duan2022factorvae} is closely related to our approach, as it uses future cross-sectional returns to infer latent factors during training. However, it is a supervised return prediction model built on engineered daily features, and its factor predictor still aggregates information across stocks at inference. In contrast, MC-JEPA learns self-supervised representations by conditioning on a market factor constructed from other assets' future target embeddings. The resulting encoder operates independently on each asset at inference.

Our key observation is that unpredictable market movements are observable during training through the realized futures of other assets. Conditioning prediction on these movements turns them from noise into supervision, shifting the learning objective from forecasting market shocks to capturing how each asset responds to them. We formalize this in a latent factor model, showing that market-conditioned prediction makes exposure necessary for optimal prediction and removes the market component from the noise against which the asset-specific signal is learned.

Building on this observation, we introduce \textbf{Market-Conditioned JEPA (MC-JEPA)}, a joint-embedding predictive architecture~\citep{assran2023self} for financial time series. A per-asset encoder embeds each asset's recent history to a representation, and a predictor estimates its future target embeddings. During training, we construct a leave-one-out market condition by aggregating the future target embeddings of other assets in the same panel. We then introduce a factor-structured predictor that incorporates this condition through loadings derived from the asset's own representation, following the structure of a single-index factor model~\citep{sharpe1964capital}. The market condition therefore cannot explain the target by itself. It helps only to the extent that the representation encodes the asset's exposure. The market condition is privileged information~\citep{duan2022factorvae} available only during training. At inference, the encoder is a standard per-asset model that requires no cross-sectional information.

We train MC-JEPA on minute-level candlesticks of the full cross-section of Chinese A-share stocks from 2015 to 2022 and evaluate the frozen representations with ridge probes on a held-out 2023--2025 period. It improves on per-asset JEPA across nine targets spanning returns, market exposure, risk, and liquidity, with the largest and most significant improvements in exposure, risk, and liquidity. It achieves the highest RankIC on all liquidity targets and on returns from two hours to five days, outperforming time-series foundation models and self-supervised baselines. In a historical trading backtest, it achieves the highest excess return and Sharpe ratio and the lowest maximum drawdown. Ablations show that the gains in return and exposure prediction depend on conditioning on the realized market movement over the same target period. Neither panel training alone nor conditioning on a different date reproduces these gains. The factor loadings of the predictor are also interpretable. Rescaled by the price volatility that per-asset normalization removes, they rank stocks by market beta nearly as well as a supervised probe, without any beta labels. A synthetic factor-model study with known parameters further shows that conditioning improves the recovery of both the true market exposure and the asset-specific component.

Our contributions are as follows:
\begin{itemize}
  \item We show under a latent factor model that market conditioning makes exposure necessary for optimal prediction and reduces noise in learning asset-specific signals.
  \item We introduce MC-JEPA, which uses market conditioning during training to learn standalone per-asset representations.
  \item We demonstrate improvements across financial prediction tasks and a trading backtest, and show that the predictor's factor loadings encode market exposure without labels.
\end{itemize}

%% file: sections/02_related_work.tex
\section{Related Work}
\label{sec:related}

\begin{figure}[t]
  \centering
  \includegraphics[width=\linewidth]{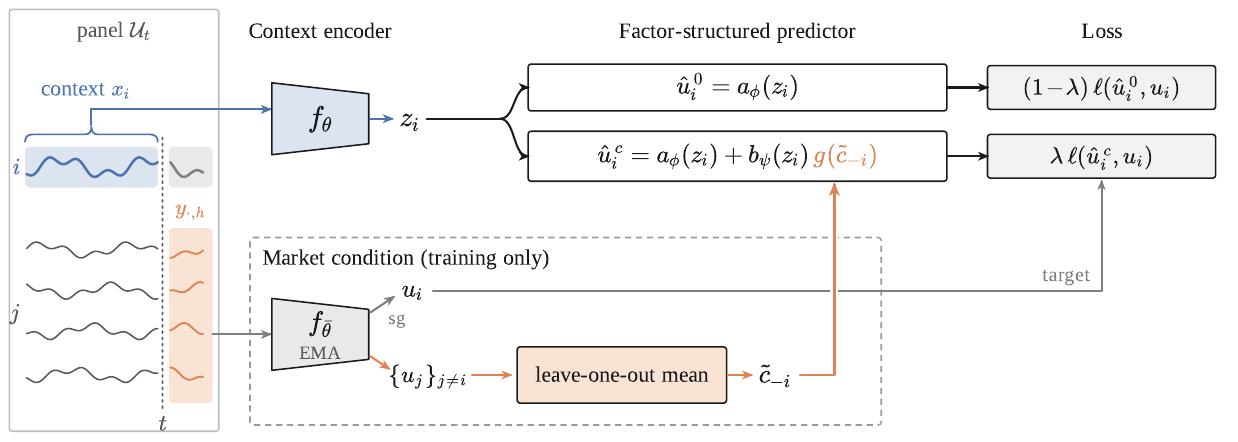}
  \caption{\textbf{MC-JEPA.} During training, each asset's future embedding is predicted both unconditionally and conditioned on a market factor derived from the future target embeddings of other assets. The factor enters through loadings computed from the asset's own representation. At inference, only the per-asset context encoder $f_\theta$ is retained.}
  \label{fig:overview}
\end{figure}

\textbf{Self-supervised learning for time series.} Self-supervised learning for time series has three main directions. Contrastive methods learn representations by aligning augmented views or temporally related windows of a series~\citep{oord2019representationlearningcontrastivepredictive, tonekaboni2021iclr-unsupervised, Ennadir2025JointEG, woo2022iclr-cost, liu2024timesurl}. Masked-reconstruction methods recover hidden input patches from their surrounding context~\citep{Yuqietal-2023-PatchTST, dong2023simmtm, pmlr-v235-goswami24a}. More recently, large-scale pretraining has produced foundation models that forecast across diverse time-series corpora~\citep{pmlr-v235-das24c, rasul2024lagllamafoundationmodelsprobabilistic, pmlr-v267-liu25an, ansari2025chronos2univariateuniversalforecasting, NEURIPS2025_ecfb69ce, ICLR2026_0c6639f4}, as well as general-purpose encoders that learn transferable representations~\citep{lin2024nutimenumericallymultiscaledembedding,fu2025frequencymaskedembeddinginferencenoncontrastive}. Joint-embedding predictive architectures (JEPA)~\citep{assran2023self} instead predict representations produced by a slowly updated target encoder, avoiding the need to reconstruct unpredictable details in input space. JEPA objectives have been adapted to time series for control~\citep{girgis2025timeseriesjepapredictiveremote}, in-context forecasting~\citep{verdenius2024latpfnjointembeddingpredictive}, general representation learning~\citep{Ennadir2025JointEG} and anomaly prediction~\citep{He2026SCJEPASL}, and recently to equity modeling~\citep{Luo2026STOCKJEPAPL, merchant2026towards}. Finance-specific foundation models such as Kronos~\citep{shi2025kronosfoundationmodellanguage} pretrain on large-scale candlestick data, yet recent evaluations find that both generic and finance-pretrained models give limited gains on financial returns~\citep{Rahimikia2025ReVisitingTS, alonso2026pretrainedtimeseriesfoundationmodels}. A common limitation of these approaches is that they encode each asset independently and predict its future from its own history. However, in financial markets, much of an asset's future variation is driven by market-wide shocks that cannot be inferred from its history alone. Consequently, such objectives provide little supervision for learning how an asset responds to these shocks. Our method addresses this limitation by retaining a per-asset JEPA encoder while conditioning its predictor on a market factor constructed from other assets' future embeddings during training, which turns unpredictable market variation into a learning signal for market exposure.

\textbf{Cross-asset dependencies in financial markets.} Asset pricing has long modeled returns through asset-specific exposures to common factors, from the CAPM~\citep{sharpe1964capital} to multi-factor models~\citep{fama1993common}. Such commonality extends beyond returns to order flow and liquidity~\citep{hasbrouck2001common}. Two lines of work are particularly relevant to ours. First, factor-based neural models learn asset-specific exposures to shared factors. Conditional autoencoders~\citep{gu2021autoencoder} model exposures as nonlinear functions of asset characteristics and combine them with factors estimated from contemporaneous realized returns. FactorVAE~\citep{duan2022factorvae} uses future cross-sectional returns as privileged information to infer latent factors during training. Both are supervised return-prediction models built on engineered daily features. In contrast, we bring the same factor exposure structure into self-supervised learning from raw time series. Asset-specific exposures are learned from each asset's representation, and the market factor is constructed from other assets' future target embeddings. Second, cross-asset neural models explicitly incorporate information from other assets through stock graphs and shared concepts~\citep{xu2021hist}, inter-stock attention with market gating~\citep{li2024master}, spatiotemporal bootstrapping over correlated series~\citep{Wang2023CorrelatedTS}, or correlation-aware adapters and group attention in foundation models~\citep{ICLR2026_ae3e1733,ansari2025chronos2univariateuniversalforecasting}. These approaches generally rely on cross-sectional context at inference, while channel-mixing architectures can be less robust to distribution shift than channel-independent ones~\citep{Han2023TheCA}. Our method instead uses the market factor only to condition the predictor during training, leaving the deployed encoder fully per-asset and independent of cross-sectional inputs.

%% file: sections/03_method.tex
\section{Method}
\label{sec:method}

We first describe a per-asset JEPA baseline (section \ref{sec:method:jepa}), then show that its objective treats unpredictable market variation as noise (section \ref{sec:method:exposure}). This motivates MC-JEPA, which conditions a factor-structured predictor on realized market variation (section \ref{sec:method:mcjepa}). Figure~\ref{fig:overview} illustrates the framework.

\subsection{Per-asset JEPA}
\label{sec:method:jepa}

\textbf{Training panels.} Each training sample consists of a decision time $t$ and a panel $\mathcal{U}_t$ of assets trading around $t$. For each asset $i\in\mathcal{U}_t$, the context $x_i\in\R^{L\times F}$ contains the most recent $L$ bars of $F$ features, ending at $t$. Each feature is standardized using the mean and standard deviation of its context window. Future bars are normalized using the same statistics, preserving their values relative to the observed history without leaking future information.

\textbf{Context and target encoders.} The context encoder $f_\theta$ divides $x_i$ into non-overlapping patches of $p$ bars~\citep{Yuqietal-2023-PatchTST}, linearly embeds each patch, and processes the sequence with a bidirectional Transformer~\citep{vaswani2017attention} equipped with learned positional embeddings and a learnable \texttt{[CLS]} token. The final \texttt{[CLS]} state gives the asset representation $z_i=f_\theta(x_i)\in\R^D$.

Following JEPA~\citep{assran2023self}, a target encoder $f_{\bar\theta}$ with the same architecture is updated by an exponential moving average, $\bar\theta\leftarrow\tau\bar\theta+(1-\tau)\theta$, with $\tau$ annealed from $\tau_0$ to $1$. For each prediction horizon $h\in\mathcal{H}$, the target block $y_{i,h}$ contains future bars beginning $h$ bars after $t$. Its embedding is
\begin{equation}
u_{i,h}=\sg\!\big(f_{\bar\theta}(y_{i,h})\big)\in\R^D,
\end{equation}
where $\sg(\cdot)$ denotes stop-gradient.

\textbf{Prediction objective.} 
A predictor $P_\phi$ takes $z_i$ and a learned horizon embedding $e_h$ to predict $u_{i,h}$. To account for differences in target scale across horizons, we standardize target embeddings using running statistics. Let $\mu_h$ and $\Sigma_h$ denote the running mean and covariance of $u_{\cdot,h}$, and define $s_h^2=\operatorname{tr}(\Sigma_h)/D$. We use the normalized mean-squared error
\begin{equation}
  \ell_h(\hat u,u)=\tfrac{1}{D}\big\|\hat u-(u-\mu_h)/s_h\big\|_2^2 .
  \label{eq:loss}
\end{equation}
Predicting the target mean yields an expected loss of approximately $1$. Thus, the normalized loss measures prediction error relative to a common mean-prediction baseline, making losses comparable across horizons.

The per-asset JEPA objective is $\mathcal{L}_{\text{JEPA}}=\E_{i,h}\,\ell_h\big(P_\phi(z_i,h),u_{i,h}\big)$. Each asset is predicted independently and the panel serves only as a training batch.

\subsection{Market Shocks: Noise for Per-Asset Prediction, Supervision under Conditioning}
\label{sec:method:exposure}

Financial assets co-move because their returns are partly driven by common market shocks~\citep{sharpe1964capital, fama1993common}. To understand what per-asset prediction learns about this dependence, consider a latent factor model of the future target embedding:
\begin{equation}
  u_{i,h}=\alpha_i+B_i f_h+\varepsilon_{i,h},
  \label{eq:factor}
\end{equation}
where $\alpha_i\in\R^D$ is an asset-specific component, $B_i\in\R^{D\times K}$ is its exposure to $K$ common factors, $f_h\in\R^K$ is the realized future market factor, and $\varepsilon_{i,h}$ is idiosyncratic noise satisfying $\E[\varepsilon_{i,h}\mid x_i,f_h]=0$. We assume $\alpha_i$ and $B_i$ are determined by the asset's history $x_i$.

Short-horizon market shocks are difficult to predict from an individual asset's history. We formalize this limiting case by assuming that $f_h$ is independent of $x_i$, with mean $m$ and covariance $\Sigma_f$. The best prediction of $u_{i,h}$ from the asset's own history is then its conditional mean
\begin{equation}
  \bar u(x_i)=\E[u_{i,h}\mid x_i]=\alpha_i+B_i m,
  \label{eq:condmean}
\end{equation}
in which the exposure $B_i$ appears only through the expected factor $m$. For a representation $z=\varphi(x_i)$, we compare the unconditional prediction error $\min_P\E\|u_{i,h}-P(z)\|^2$ with the market-conditioned prediction error $\min_P\E\|u_{i,h}-P(z,f_h)\|^2$, where the predictor additionally observes the realized factor.

\begin{proposition}
\label{prop:identify}
Under~\eqref{eq:factor} and the assumptions above, let $z=\varphi(x_i)$ be a representation.

\textup{(a)} Over all representations, the minimum unconditional prediction error is $\E\|B_i(f_h-m)\|^2+\E\|\varepsilon_{i,h}\|^2$. A representation $z$ attains it if and only if $\bar u(x_i)$ can be recovered from $z$.

\textup{(b)} If $\Sigma_f\succ0$, the minimum market-conditioned prediction error over all representations is $\E\|\varepsilon_{i,h}\|^2$. A representation $z$ attains it if and only if both $\alpha_i$ and $B_i$ can be recovered from $z$.
\end{proposition}

The proof is in Appendix~\ref{app:proof}. Part (a) shows that unconditional prediction requires only the conditional mean $\bar u(x_i)$, not the exposure $B_i$ separately. The unpredictable component $B_i(f_h-m)$ contributes to irreducible error without providing additional supervision for exposure. Although market factors may be partially predictable in practice, their unpredictable variation remains uninformative for learning exposure under unconditional prediction.

Part (b) shows how conditioning changes this. When the predictor observes the realized market factor, its prediction error includes an additional term that penalizes exposure information missing from the representation. Achieving the minimum error requires preserving $B_i$. Therefore, conditioning turns irreducible market variation into supervision for learning asset-specific exposure.

\textbf{Conditioning reduces prediction noise.}
By accounting for realized market shocks, conditioning reduces the noise in learning asset-specific signals. This complements exposure learning by improving beta and risk prediction through market exposure and benefiting return and volume prediction through cleaner asset-specific representations. In practice, the learned condition reduces unconditional prediction error by 4.5\% at $K=1$, indicating a modest denoising effect (Appendix~\ref{app:denoise}).

\subsection{MC-JEPA}
\label{sec:method:mcjepa}

Motivated by Proposition~\ref{prop:identify}, we introduce a market condition estimated from other assets' future embeddings and a factor-structured predictor that uses this condition through loadings computed from the asset's own representation.

\textbf{Market condition from future embeddings.} Although the realized market factor $f_h$ is latent, its common component is reflected in the future embeddings of other assets. For each scored asset $i$, we construct a leave-one-out market condition from a pool $\mathcal{M}_t\subseteq\mathcal{U}_t$,
\begin{equation}
  c_{-i,h}=\frac{1}{|\mathcal{M}_t|-1}\sum_{j\in\mathcal{M}_t\setminus\{i\}}u_{j,h},
  \qquad
  \tilde c_{-i,h}=(c_{-i,h}-\mu^c_h)/s^c_h,
  \label{eq:condition}
\end{equation}
where $\mu^c_h$ and $s^c_h$ are running statistics of the full-pool mean, computed as in~\eqref{eq:loss}. When idiosyncratic noise is weakly correlated across assets, averaging reduces this noise while retaining variation driven by common market factors. As long as the factor effects do not cancel across assets, $c_{-i,h}$ serves as a noisy proxy for the realized market factor $f_h$.

We exclude asset $i$ from the mean so that the condition never contains its own target, which would allow the predictor to copy the answer. The condition is built from target embeddings, so it carries no gradient and costs only forward passes of the target encoder. The condition pool can also be larger than the scored asset set $\mathcal{S}_t\subseteq\mathcal{M}_t$, allowing the market estimate to aggregate more assets without increasing the number of online encoder evaluations.

\textbf{Factor-structured predictor.} To turn the market condition into supervision for asset-specific exposure, we structure the predictor so that the market factor influences each asset's prediction through loadings inferred from its own representation. Following the exposure--factor decomposition in~\eqref{eq:factor}, we produce an unconditional and a market-conditioned prediction:
\begin{equation}
  \hat u^{\,0}_{i,h}=a_\phi(z_i,h),
  \qquad
  \hat u^{\,c}_{i,h}=a_\phi(z_i,h)+B_\psi(z_i,h)\,g(\tilde c_{-i,h}),
  \label{eq:predictor}
\end{equation}
where $a_\phi$ is a base predictor, $g$ extracts a low-dimensional market factor, and $B_\psi$ computes asset-specific loadings from $z_i$.

This factorization restricts how market information enters prediction. Both passes share the same base $a_\phi$, which never observes the condition. Any asset-dependent adjustment from the realized market factor must be mediated by loadings computed from $z_i$. The bilinear term $B_\psi(z_i,h)g(\tilde c_{-i,h})$ mirrors the latent factor model. When the learned factor retains the relevant market variation, accurate conditional prediction rewards representations that encode the asset's exposure. The factor dimension $K$ further limits the capacity of the market-dependent adjustment.

We use a single factor ($K=1$), so $g(\tilde c_{-i,h})=w^\top\tilde c_{-i,h}$ is a scalar and $B_\psi(z_i,h)$ reduces to a loading vector $b_\psi(z_i,h)\in\R^D$. This gives a latent analogue of the single-index model~\citep{sharpe1964capital}, in which each asset's response to a common factor is determined by its own representation.

\textbf{Training objective.} We supervise both predictions against the same future target:
\begin{equation}
  \mathcal{L}=(1-\lambda)\,\E_{i,h}\,\ell_h\big(\hat u^{\,0}_{i,h},u_{i,h}\big)+\lambda\,\E_{i,h}\,\ell_h\big(\hat u^{\,c}_{i,h},u_{i,h}\big),
  \label{eq:objective}
\end{equation}
where the expectation is over decision times $t$, scored assets $i\in\mathcal{S}_t$ and horizons $h\in\mathcal{H}$. The weight $\lambda\in[0,1]$ controls the contribution of the conditional objective and $\lambda=0$ recovers the per-asset JEPA baseline.

The unconditional term preserves the representation's ability to predict what is inferable from the asset's own history, while the conditional term additionally supervises its response to realized market variation. Both passes share the encoder outputs and base prediction, so the overhead over per-asset JEPA is the loading network and the target encoder's forward passes over the condition pool.

\textbf{Inference.} The market condition is privileged information~\citep{duan2022factorvae} during training only. Downstream tasks use $z_i=f_\theta(x_i)$, which depends on asset $i$'s own history alone. No peer assets, future observations, or market index are required at inference. MC-JEPA thus incorporates cross-asset supervision during training while retaining the inference interface and cost of a standard per-asset encoder.

%% file: sections/04_experiments.tex
\section{Experiments}
\label{sec:experiments}

\subsection{Experimental Setup}

\textbf{Datasets and Baselines.} We use 1-minute OHLCV bars of Chinese A-share stocks from 2015 to 2025. All splits are chronological. Pretraining uses decision times before July 2022. For evaluation we freeze each encoder and embed one window per stock and trading day, ending at the target's decision time. Daily targets are decided at the 15:00 close. Minute targets are decided at the 11:30 morning close. We fit ridge probes on 2019-01 to 2022-06, select the ridge penalty and the hyperparameters of MC-JEPA on 2022-07 to 2022-12, and report results on 2023-01 to 2025-12. We compare against three groups of baselines. \emph{(i) Finance-specific baselines.} We use the Alpha158 daily factors~\citep{Yang2020QlibAA}. Like the encoders, it is read out with a ridge probe. \emph{(ii) Pretrained time-series foundation models.} We use frozen Kronos~\citep{shi2025kronosfoundationmodellanguage}, TimesFM-2.5~\citep{pmlr-v235-das24c}, Chronos-2~\citep{ansari2025chronos2univariateuniversalforecasting} and MOMENT~\citep{pmlr-v235-goswami24a}. \emph{(iii) Self-supervised objectives.} We compare objectives under the same encoder architecture, training data and compute budget. These include masked patch reconstruction (Masked recon.)~\citep{Yuqietal-2023-PatchTST, dong2023simmtm}, next-patch autoregression (Next-patch AR), TS2Vec~\citep{yue2022ts2vec}, CPC~\citep{oord2019representationlearningcontrastivepredictive}, and per-asset JEPA (section \ref{sec:method:jepa}). Ablations that isolate each component of our method are in section \ref{sec:ablation}. Each encoder is read out from the \texttt{[CLS]} state if its objective trains it (JEPA variants and CPC), and from the mean of its patch states otherwise. For every baseline we checked that this is the better of the two readouts. Every encoder we train is run with three seeds, and we report the mean and standard deviation.

\textbf{Evaluation metrics.} We evaluate nine downstream targets spanning returns, market exposure, risk, and liquidity. These include future returns over horizons from 120 minutes to 20 days, forward 20-day market beta, realized volatility, relative volume, and the fraction of next-day volume traded at the open. Our primary metric is RankIC, computed as the Spearman correlation between predictions and realized targets across stocks on each test day, averaged over the test period. We further assess economic value through a historical long-only backtest using the Ret 20d predictions, reporting annualized excess return (AER), Sharpe ratio, information ratio (IR), and maximum drawdown (MDD). Appendix~\ref{app:metrics} provides detailed definitions of all targets and metrics.

\textbf{Implementation details.} Inputs are 1-minute OHLCV bars with a context of $L=960$ bars, split into 96 patches of $p=10$ bars. The encoder is a 12-layer Transformer with width $D=256$ (about 10M parameters). Targets are 30-bar blocks at horizons $\mathcal{H}=\{1,60,240\}$ bars, spanning the next half hour to the next trading day. Each step samples 128 decision times, each with a condition pool of $|\mathcal{M}_t|=32$ assets of which $|\mathcal{S}_t|=8$ are scored. The pool is drawn uniformly at random from all assets around $t$, and the scored assets are a uniform random subset of the pool. Appendix~\ref{app:hparams} lists all hyperparameters.

\begin{table}[t]
\centering
\caption{Main results on the 2023--2025 test set. We report RankIC of ridge probes fitted to frozen encoder representations or Alpha158 features. Encoders trained in our experiments report mean$_{\pm\mathrm{std}}$ over three seeds. \textbf{Bold} and \underline{underlined} values indicate the best and second-best scores, respectively.}
\label{tab:main}
\newcommand{\s}[1]{$_{\pm#1}$}
\setlength{\tabcolsep}{3pt}
\resizebox{\textwidth}{!}{%
\begin{tabular}{l cccc c c ccc}
\toprule
 & \multicolumn{4}{c}{Return} & Exposure & Risk & \multicolumn{3}{c}{Liquidity} \\
\cmidrule(lr){2-5}\cmidrule(lr){6-6}\cmidrule(lr){7-7}\cmidrule(lr){8-10}
Method & Ret 120m & Ret 1d & Ret 5d & Ret 20d & Beta 20d & Vol 120m & Volume 120m & Volume 1d & Open vol share \\
\midrule
Alpha158 & .0425 & .0591 & .0867 & .1011 & .356 & .364 & .257 & .417 & .353 \\
\midrule
Kronos-mini & .0521 & .0652 & .1028 & .1395 & .324 & .397 & .466 & .367 & .358 \\
Kronos-small & .0601 & .0673 & .1035 & \underline{.1418} & .333 & .419 & .552 & .423 & .376 \\
Kronos-base & .0647 & .0649 & .1048 & .1409 & .348 & .450 & .568 & .433 & .381 \\
TimesFM-2.5 & .0564 & .0626 & .0983 & .1358 & .350 & .424 & .515 & .406 & .373 \\
Chronos-2 & .0649 & .0639 & .1023 & .1390 & .362 & .491 & \underline{.580} & \underline{.440} & .387 \\
MOMENT-small & .0552 & .0677 & .0972 & .1261 & .293 & .333 & .527 & .390 & .358 \\
MOMENT-base & .0564 & .0643 & .0962 & .1262 & .294 & .324 & .524 & .387 & .356 \\
\midrule
Masked recon. & .0605\s{.0041} & .0682\s{.0008} & .0998\s{.0010} & .1295\s{.0013} & .324\s{.017} & .317\s{.007} & .492\s{.016} & .393\s{.005} & .361\s{.008} \\
Next-patch AR & .0648\s{.0026} & .0681\s{.0004} & .1002\s{.0010} & .1315\s{.0017} & \underline{.382}\s{.006} & .395\s{.009} & .540\s{.002} & .419\s{.001} & .390\s{.001} \\
TS2Vec & .0586\s{.0018} & .0687\s{.0005} & \underline{.1058}\s{.0010} & \textbf{.1454}\s{.0012} & \underline{.382}\s{.002} & .384\s{.017} & .469\s{.007} & .377\s{.005} & .364\s{.002} \\
CPC & .0634\s{.0024} & .0668\s{.0029} & .1011\s{.0049} & .1397\s{.0008} & \textbf{.391}\s{.004} & \textbf{.552}\s{.005} & .542\s{.004} & .401\s{.005} & .390\s{.002} \\
Per-asset JEPA & \underline{.0677}\s{.0026} & \underline{.0712}\s{.0009} & .1038\s{.0006} & .1337\s{.0025} & .359\s{.003} & .444\s{.017} & .573\s{.005} & \underline{.440}\s{.003} & \underline{.391}\s{.002} \\
\midrule
MC-JEPA & \textbf{.0694}\s{.0009} & \textbf{.0746}\s{.0003} & \textbf{.1074}\s{.0004} & .1389\s{.0015} & .374\s{.003} & \underline{.517}\s{.007} & \textbf{.594}\s{.006} & \textbf{.454}\s{.005} & \textbf{.397}\s{.001} \\
\bottomrule
\end{tabular}}
\end{table}

\begin{table}[t]
\centering
\caption{Backtest results on the 2023--2025 test period. Each day, the strategy selects the 7--8 stocks with the highest ridge-predicted 20-day returns among point-in-time CSI300 index members and holds them for 20 trading days. Results are net of 5/15\,bp transaction costs, with excess returns measured against the CSI300 index. AER denotes annualized excess return, Sharpe the annualized Sharpe ratio, IR the annualized information ratio of the excess return, and MDD the magnitude of maximum drawdown. Encoders trained in our experiments report mean$_{\pm\mathrm{std}}$ over three seeds. \textbf{Bold} and \underline{underlined} values indicate the best and second-best scores, respectively.}
\label{tab:backtest}
\newcommand{\s}[1]{$_{\pm#1}$}
\setlength{\tabcolsep}{4pt}
\begin{tabular}{l cccc}
\toprule
Method & AER $\uparrow$ & Sharpe $\uparrow$ & IR $\uparrow$ & MDD $\downarrow$ \\
\midrule
Alpha158 & 0.103 & 1.07 & 1.16 & 0.135 \\
\midrule
Kronos-mini & \underline{0.128} & 1.09 & \textbf{1.32} & 0.159 \\
Kronos-small & 0.101 & 0.95 & 1.12 & 0.177 \\
Kronos-base & 0.107 & 0.99 & 1.16 & 0.161 \\
TimesFM-2.5 & 0.096 & 0.89 & 1.07 & 0.190 \\
Chronos-2 & 0.092 & 0.86 & 1.01 & 0.186 \\
MOMENT-small & 0.044 & 0.61 & 0.57 & 0.182 \\
MOMENT-base & 0.069 & 0.77 & 0.86 & 0.197 \\
\midrule
Masked recon. & 0.118\s{0.017} & 1.08\s{0.12} & \underline{1.24}\s{0.15} & 0.151\s{0.010} \\
Next-patch AR & 0.083\s{0.009} & 0.90\s{0.05} & 0.85\s{0.08} & 0.154\s{0.007} \\
TS2Vec & 0.090\s{0.012} & 0.85\s{0.07} & 0.97\s{0.13} & 0.184\s{0.013} \\
CPC & 0.030\s{0.007} & 0.50\s{0.05} & 0.36\s{0.09} & 0.247\s{0.014} \\
Per-asset JEPA & 0.117\s{0.006} & \underline{1.11}\s{0.03} & 1.20\s{0.07} & \underline{0.132}\s{0.010} \\
\midrule
MC-JEPA & \textbf{0.135}\s{0.011} & \textbf{1.19}\s{0.08} & \textbf{1.32}\s{0.09} & \textbf{0.117}\s{0.001} \\
\bottomrule
\end{tabular}
\end{table}

\begin{table}[t]
\centering
\caption{Ablation results on the 2023--2025 test period. We report RankIC of ridge probes fitted to frozen representations, using the targets in Table~\ref{tab:main}. Each variant changes only the indicated component or hyperparameter relative to MC-JEPA. Results are reported as mean$_{\pm\mathrm{std}}$ over three seeds. \textbf{Bold} values indicate the best score in each column.}
\label{tab:ablation}
\newcommand{\s}[1]{$_{\pm#1}$}
\setlength{\tabcolsep}{3pt}
\resizebox{\textwidth}{!}{%
\begin{tabular}{l cccc c c ccc}
\toprule
 & \multicolumn{4}{c}{Return} & Exposure & Risk & \multicolumn{3}{c}{Liquidity} \\
\cmidrule(lr){2-5}\cmidrule(lr){6-6}\cmidrule(lr){7-7}\cmidrule(lr){8-10}
Variant & Ret 120m & Ret 1d & Ret 5d & Ret 20d & Beta 20d & Vol 120m & Volume 120m & Volume 1d & Open vol share \\
\midrule
\multicolumn{10}{l}{\emph{(a) Components}} \\
Per-asset JEPA & .0677\s{.0026} & .0712\s{.0009} & .1038\s{.0006} & .1337\s{.0025} & .359\s{.003} & .444\s{.017} & .573\s{.005} & .440\s{.003} & .391\s{.002} \\
Panel JEPA ($\lambda=0$) & .0663\s{.0013} & .0714\s{.0004} & .1042\s{.0010} & .1354\s{.0006} & .345\s{.005} & .439\s{.029} & .572\s{.004} & .439\s{.003} & .388\s{.001} \\
Placebo condition (other date) & .0667\s{.0044} & .0720\s{.0015} & .1031\s{.0016} & .1350\s{.0036} & .362\s{.002} & .487\s{.023} & .584\s{.003} & .448\s{.004} & .392\s{.003} \\
Unfactored predictor & .0658\s{.0006} & .0732\s{.0015} & .1066\s{.0011} & .1365\s{.0013} & .366\s{.002} & .498\s{.016} & .589\s{.006} & .450\s{.004} & .394\s{.001} \\
\midrule
\multicolumn{10}{l}{\emph{(b) Condition weight $\lambda$}} \\
$\lambda=0.25$ & .0663\s{.0014} & .0719\s{.0004} & .1040\s{.0007} & .1350\s{.0004} & .365\s{.002} & .505\s{.003} & .589\s{.001} & .450\s{.003} & .396\s{.001} \\
$\lambda=0.5$ & .0685\s{.0020} & .0735\s{.0014} & .1055\s{.0025} & .1382\s{.0027} & .367\s{.002} & .519\s{.012} & .589\s{.006} & .450\s{.003} & .396\s{.001} \\
$\lambda=1$ & .0674\s{.0031} & .0742\s{.0014} & .1059\s{.0014} & .1375\s{.0016} & .371\s{.005} & .524\s{.007} & .594\s{.005} & .453\s{.001} & .397\s{.001} \\
\midrule
\multicolumn{10}{l}{\emph{(c) Number of factors $K$}} \\
$K=2$ & .0677\s{.0004} & .0720\s{.0006} & .1050\s{.0008} & .1379\s{.0017} & .378\s{.007} & .527\s{.010} & .598\s{.004} & .454\s{.005} & \textbf{.399}\s{.003} \\
$K=4$ & \textbf{.0702}\s{.0010} & .0713\s{.0021} & .1056\s{.0023} & .1380\s{.0015} & .379\s{.004} & \textbf{.540}\s{.017} & .600\s{.002} & .455\s{.001} & \textbf{.399}\s{.001} \\
$K=8$ & .0673\s{.0027} & .0730\s{.0012} & .1058\s{.0008} & .1371\s{.0009} & .375\s{.002} & .524\s{.007} & .598\s{.003} & .456\s{.003} & .398\s{.002} \\
$K=16$ & .0682\s{.0025} & .0720\s{.0022} & .1056\s{.0012} & .1383\s{.0009} & .377\s{.000} & .538\s{.004} & .599\s{.001} & .453\s{.002} & .398\s{.001} \\
$K=32$ & .0680\s{.0007} & .0712\s{.0017} & .1050\s{.0013} & .1379\s{.0014} & \textbf{.382}\s{.005} & \textbf{.540}\s{.002} & .599\s{.001} & .455\s{.002} & \textbf{.399}\s{.000} \\
$K=64$ & .0689\s{.0011} & .0715\s{.0013} & .1048\s{.0007} & .1372\s{.0003} & .376\s{.001} & .536\s{.012} & \textbf{.603}\s{.004} & \textbf{.458}\s{.002} & \textbf{.399}\s{.001} \\
$K=128$ & .0684\s{.0004} & .0712\s{.0020} & .1045\s{.0022} & .1368\s{.0021} & .377\s{.001} & .538\s{.013} & .602\s{.002} & .457\s{.000} & \textbf{.399}\s{.002} \\
\midrule
\multicolumn{10}{l}{\emph{(d) Condition pool size $|\mathcal{M}_t|$}} \\
$|\mathcal{M}_t|=8$ & .0675\s{.0017} & .0727\s{.0008} & .1049\s{.0018} & .1365\s{.0019} & .369\s{.001} & .509\s{.008} & .593\s{.001} & .451\s{.003} & .396\s{.001} \\
$|\mathcal{M}_t|=16$ & .0679\s{.0014} & .0743\s{.0007} & .1058\s{.0001} & .1373\s{.0010} & .371\s{.003} & .508\s{.022} & .593\s{.004} & .451\s{.001} & .396\s{.001} \\
$|\mathcal{M}_t|=64$ & .0670\s{.0020} & .0728\s{.0007} & .1055\s{.0003} & .1376\s{.0009} & .368\s{.002} & .518\s{.008} & .591\s{.003} & .452\s{.004} & .397\s{.000} \\
$|\mathcal{M}_t|=128$ & .0685\s{.0012} & .0737\s{.0017} & .1058\s{.0015} & .1385\s{.0016} & .370\s{.007} & .524\s{.003} & .593\s{.005} & .453\s{.004} & .395\s{.001} \\
\midrule
\begin{tabular}[t]{@{}l@{}}MC-JEPA\\($\lambda=0.75$, $K=1$, $|\mathcal{M}_t|=32$)\end{tabular} & .0694\s{.0009} & \textbf{.0746}\s{.0003} & \textbf{.1074}\s{.0004} & \textbf{.1389}\s{.0015} & .374\s{.003} & .517\s{.007} & .594\s{.006} & .454\s{.005} & .397\s{.001} \\
\bottomrule
\end{tabular}}
\end{table}

\subsection{Results}
\label{sec:results}

Table~\ref{tab:main} reports results on nine downstream targets. Tables~\ref{tab:minute-horizons} and~\ref{tab:daily-extra} in Appendix report the minute targets at shorter horizons and four more daily targets.

\textbf{One representation for return, risk and liquidity.} MC-JEPA achieves the best RankIC on six of the nine targets, covering returns from two hours to five days and all three liquidity targets, and is second on intraday volatility, behind CPC. On Ret 20d, TS2Vec leads with a RankIC of 0.1454, compared with 0.1389 for MC-JEPA. Among the trained encoders, MC-JEPA has the smallest seed standard deviation on the three return targets up to five days.

\textbf{Conditioning improves on per-asset JEPA.} Per-asset JEPA uses the same setup and differs only in the conditional term of~\eqref{eq:objective}. Adding the condition improves all nine targets. Paired Newey--West tests (Table~\ref{tab:paired}) show the gain significant at the 5\% level on seven of the nine targets. The gains are not uniform across target types. Gains on exposure, risk, and liquidity are all significant, with $t$-statistics ranging from 6.9 to 49.4. Return gains are smaller and significant only for Ret 1d and Ret 20d. Next-patch AR, TS2Vec and CPC score higher on Beta 20d but are clearly worse on returns and volume. We compare the representation with trailing-beta estimators in Appendix~\ref{app:exposure}.

\textbf{Foundation models do not close the gap.} Frozen time-series foundation models are competitive on liquidity, where Chronos-2 is second on both volume targets. They are weaker on returns. Every TSFM is below MC-JEPA on all three return horizons. Scale does not change this. Moving from Kronos-mini to Kronos-base improves volatility and volume, but Ret 1d stays between 0.065 and 0.067.

\textbf{Backtest.} Table~\ref{tab:backtest} turns the Ret 20d probe into a long-only CSI300 strategy. MC-JEPA has the highest annualized excess return, the highest Sharpe ratio and the smallest maximum drawdown. Compared with per-asset JEPA, it raises excess return by 1.8 points and lowers drawdown from 0.132 to 0.117, with a seed standard deviation of 0.001 on drawdown. Figure~\ref{fig:backtest-curve} shows the cumulative excess return of each method.

\subsection{Ablations}
\label{sec:ablation}

Table~\ref{tab:ablation} isolates the contribution of each component of MC-JEPA. Tables~\ref{tab:ablation-minute} and~\ref{tab:ablation-daily} in Appendix~\ref{app:results} report the same variants on the additional targets.

\textbf{The gain comes from the aligned conditioning.} Panel JEPA removes conditioning by setting $\lambda=0$ while keeping the rest of the setup unchanged. It performs comparably to per-asset JEPA on returns and liquidity but worse on Beta 20d, indicating that panel sampling alone does not explain the gains. The placebo retains the full method but constructs the condition in~\eqref{eq:condition} from a different decision time, breaking its alignment with the realized market movement over the target block. It recovers some of the gains on volatility and volume but not in returns or market exposure. Paired Newey--West tests (Table~\ref{tab:paired}) confirm that MC-JEPA is significantly better than the placebo on all nine targets of Table~\ref{tab:main} at the 5\% level. These results suggest that the return and exposure gains depend on conditioning on the realized market movement over the same target block, consistent with Proposition~\ref{prop:identify}.

\textbf{Factor structure.} The unfactored predictor feeds the condition to a predictor that reads it jointly with $z_i$, so market information can shape the prediction without passing through asset-specific loadings. It falls below MC-JEPA on all nine targets of Table~\ref{tab:main}, indicating that routing the condition through loadings computed from $z_i$ is necessary.

\textbf{Condition weight.} Every $\lambda>0$ improves market exposure, risk and liquidity prediction over $\lambda=0$, while returns prediction benefits only when $\lambda\ge0.5$. Improvements on liquidity saturate at $\lambda=0.25$, while Beta 20d and Vol 120m keep improving as $\lambda$ grows. We select $\lambda$ by return prediction on the validation period, where Ret 1d and Ret 5d peak at $\lambda=0.75$ (Table~\ref{tab:ablation-val}), and they also peak there on the test period. Removing the unconditional term entirely yields the best Vol 120m performance but slightly degrades return prediction.

\textbf{Number of factors.} Every $K>1$ improves market exposure and risk relative to $K=1$, though not monotonically in $K$. At $K=32$, Beta 20d reaches 0.382, matching next-patch AR and TS2Vec in Table~\ref{tab:main}. Return prediction goes the other way, with $K=1$ performing best on Ret 1d, Ret 5d and Ret 20d. Since we select $K$ on the validation period, we adopt $K=1$ to prioritize return prediction while retaining a direct latent analogue of the single-index model. These results suggest that additional factors capture cross-sectional structure relevant to risk and exposure, allowing $K$ to be selected according to the downstream task.

\textbf{Condition pool size.} Return and exposure prediction peak at the default pool of $|\mathcal{M}_t|=32$. Both restricting the pool to the scored assets and widening it beyond 32 lower Ret 1d, Ret 5d, Ret 20d and Beta 20d. Vol 120m is the exception, improving with pool size overall, consistent with a less noisy market estimate, while the liquidity targets barely change. A wider pool helps the risk target, but not return and exposure prediction.

\subsection{Market Exposure Analysis}
\label{sec:analysis}

\textbf{The factor loadings encode market exposure without labels.} The loadings of the factor-structured predictor can be read as an exposure estimate without beta labels. We project each stock's loading onto the mean loading of its date, and multiply by the coefficient of variation of the stock's close price over its context window, which is the scale removed by the per-asset normalization. Its RankIC with Beta 20d is 0.368, compared with 0.374 for a supervised ridge probe on the representation and 0.287 for price scale alone. Without the scale, the projection has a RankIC of $-0.06$. This is because normalization hides each asset's volatility, so the loadings capture its correlation with the market rather than its beta. Consistent with this, the projection predicts each stock's realized correlation with the market over the next 20 days beyond its trailing four-day correlation and volatility, with a partial RankIC of 0.105. A detailed discussion is provided in Appendix~\ref{app:exposure}.

\subsection{Synthetic Study}
\label{sec:analysis:synthetic}

Real data offers only one level of market predictability and an unknown factor structure. We simulate a panel with both under control to examine whether conditioning improves the representation of market exposure $B_i$ and the asset-specific component $a_i$, as discussed in Section~\ref{sec:method:exposure}.

\textbf{Setup.} Returns follow a single-factor model, $r_{i,t}=a_i+B_if_t+e_{i,t}$, with random $a_i$ and $B_i$. We control how predictable the next factor value $f_{t+1}$ is from the past, from $R^2=0$ to $R^2=0.9$. A small MLP encoder sees each asset's recent returns and the factor path, and is trained with per-asset JEPA or with MC-JEPA. We then fit linear probes for $B_i$ and $a_i$ on the frozen representation of new assets. Details are in Appendix~\ref{app:synthetic}.

\begin{figure}[t]
\centering
\includegraphics[width=\linewidth]{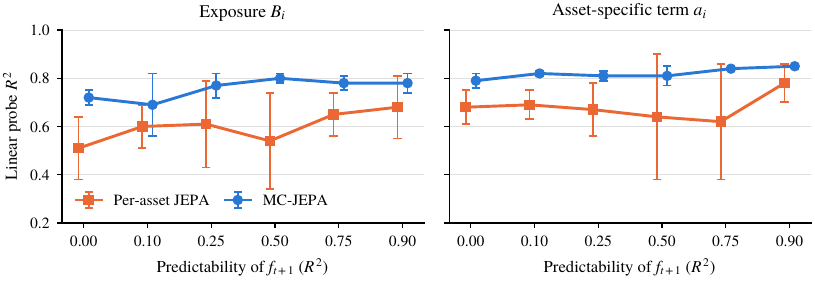}
\caption{Synthetic factor-model results. We report the $R^2$ of linear probes fitted to frozen representations for market exposure $B_i$ (left) and the asset-specific component $a_i$ (right), evaluated on held-out assets. The x-axis varies the predictability of the next market factor, measured by the $R^2$ of predicting $f_{t+1}$ from its history. Points show the mean and error bars the standard deviation over five seeds.}
\label{fig:synthetic}
\end{figure}

\textbf{Results.} Figure~\ref{fig:synthetic} shows that MC-JEPA improves the recovery of both quantities at every level of market-factor predictability. For the exposure $B_i$, linear-probe $R^2$ increases from 0.51--0.68 with per-asset JEPA to 0.69--0.80 with MC-JEPA, with less variation across seeds. Per-asset JEPA learns part of $B_i$ from how the asset co-moves with the factor inside the input window, and this signal does not depend on whether $f_{t+1}$ is predictable. Even so, conditioning recovers the exposure better. MC-JEPA also recovers the asset-specific component $a_i$ more accurately, improving $R^2$ from 0.62--0.78 to 0.79--0.85. This is consistent with the denoising effect, as conditioning explains away the market noise $B_if_{t+1}$ that otherwise interferes with learning $a_i$. The learned condition removes 30\% of the unconditional error when $f_{t+1}$ is unpredictable and 5\% at $R^2=0.9$ (Appendix~\ref{app:denoise}). The gain on $a_i$ persists at $R^2=0.9$, so denoising does not account for all of it. This second effect is the one relevant to return prediction on real data, where the probe ranks assets without access to the realized market shock.

%% file: sections/06_conclusion.tex
\section{Conclusion}
\label{sec:conclusion}

We introduced MC-JEPA, a self-supervised method that turns unpredictable market movements into supervision. Conditioning on the realized market factor makes market exposure necessary for optimal prediction and denoises the asset-specific signal. During training, a factor-structured predictor conditions on a leave-one-out market signal constructed from other assets' future target embeddings, with loadings computed from the asset's own representation. The condition is used only during training, leaving a per-asset encoder that requires no cross-sectional information at inference. On 11 years of minute-level Chinese A-share data, MC-JEPA improves over per-asset JEPA on all nine targets spanning returns, market exposure, risk, and liquidity, most strongly on exposure, risk, and liquidity. It achieves the highest RankIC on all liquidity targets and on returns from two hours to five days, as well as the best performance in a historical backtest. Ablations show that the gains in return and exposure prediction depend on aligning the market condition with the realized market movement over the target period. A synthetic factor-model study further demonstrates that conditioning improves the recovery of both the true market exposure and the asset-specific component.

%% file: sections/07_limitations.tex
\section{Limitations}
\label{sec:limitations}

All real-data experiments are conducted on Chinese A-share equities at a single bar frequency. Market-specific characteristics, including heavy retail participation, daily price limits, and T+1 settlement, may influence the impact of market-wide shocks on individual stocks. Whether our gains generalize to other markets, asset classes, or frequencies remains untested. Furthermore, Proposition~\ref{prop:identify} assumes a linear factor model with access to the true factor, whereas MC-JEPA relies on an estimated leave-one-out market condition. Our use of a single factor ($K=1$) also leaves richer cross-sectional structure, such as industry and style factors, to be captured implicitly. Because inputs and targets are standardized per window, the learned loadings capture co-movement with the market rather than beta. Multiplying them by the window's price volatility recovers a cross-sectional beta ranking. Targets that preserve each asset's scale could make the loadings a beta directly.

%% file: sections/A_appendix.tex
\section{Proof of Proposition~\ref{prop:identify}}
\label{sec:appendix}
\label{app:proof}
We omit the asset and horizon indices and consider the factor model
\[
  u=\alpha+Bf+\varepsilon,
\]
where $\alpha=\alpha(x)$ and $B=B(x)$ are determined by the asset history $x$. Let $z=\varphi(x)$ be its learned representation. We assume finite second moments, $f$ is independent of $x$, with $\E[f]=m$ and $\operatorname{Cov}(f)=\Sigma_f$, and $\E[\varepsilon\mid x,f]=0$.

Recall the conditional mean $\bar u(x)=\E[u\mid x]=\alpha+Bm$ from~\eqref{eq:condmean}, the optimal squared-error prediction given the asset's history alone. We show that unconditional prediction requires only $\bar u(x)$, whereas conditioning on $f$ additionally requires the exposure $B$.

\textbf{(a) Unconditional prediction.} Since $z$ is a function of $x$, the orthogonality of conditional expectations gives
\begin{align}
  \E\|u-P(z)\|^2
  &=\E\|u-\bar u(x)\|^2
    +\E\|\bar u(x)-P(z)\|^2 \nonumber\\
  &=\E\|\varepsilon\|^2
    +\E\|\bar u(x)-P(z)\|^2
    +\E\|B(f-m)\|^2.
\end{align}
The first and last terms are irreducible errors arising from unpredictable market variation and idiosyncratic noise. Neither depends on the representation or predictor. The middle term measures how much of the conditional mean $\bar u(x)$ cannot be recovered from $z$. For a fixed $z$, it is minimized by $P(z)=\E[\bar u(x)\mid z]$, which gives
\[
  \min_P\E\|u-P(z)\|^2
  =\E\|B(f-m)\|^2+\E\|\varepsilon\|^2+\E\|\delta_{\bar u}\|^2,
  \qquad
  \delta_{\bar u}=\bar u(x)-\E[\bar u(x)\mid z].
\]

Consequently, the minimum prediction error equals the floor $\E\|B(f-m)\|^2+\E\|\varepsilon\|^2$ if and only if $\delta_{\bar u}=0$ almost surely, that is, $\bar u(x)$ can be recovered from $z$. The representation need not preserve $B$ separately, since exposure to unpredictable market variation contributes only to the irreducible error and provides no additional supervision.

\textbf{(b) Market-conditioned prediction.} When the predictor also observes the realized factor $f$, the optimal prediction given $(x,f)$ becomes
\[
  \E[u\mid x,f]
  =\alpha+Bf
  =\bar u(x)+B(f-m).
\]
Unlike the unconditional mean, this prediction depends on the asset's exposure to the realized factor variation.

For a fixed representation $z$, the optimal predictor is
\begin{align}
  P^\star(z,f)
  &=\E[u\mid z,f] =\E[\bar u(x)\mid z]
    +\E[B\mid z](f-m),
\end{align}
where the second equality follows from the independence of $f$ and $x$.

Define the parts of the conditional mean and exposure that are not recoverable from $z$:
\[
  \delta_{\bar u}=\bar u(x)-\E[\bar u(x)\mid z],
  \qquad
  \delta_B=B-\E[B\mid z].
\]
The prediction error can then be written as
\[
  u-P^\star(z,f)
  =\varepsilon+\delta_{\bar u}+\delta_B(f-m).
\]
Since $\E[\varepsilon\mid x,f]=0$ and $f-m$ is independent of $(\delta_{\bar u},\delta_B)$ with zero mean, all cross terms vanish. Therefore,
\begin{align}
  \E\|u-P^\star(z,f)\|^2
  &=\E\|\varepsilon\|^2
    +\E\|\delta_{\bar u}+\delta_B(f-m)\|^2 \nonumber\\
  &=\E\|\varepsilon\|^2+\E\|\delta_{\bar u}\|^2
    +\E\operatorname{tr}(\delta_B\Sigma_f\delta_B^\top).
\end{align}
The last term follows from
\[
  \E\|\delta_B(f-m)\|^2
  =\E\operatorname{tr}
    (\delta_B\Sigma_f\delta_B^\top).
\]

The decomposition shows that conditioning introduces an additional penalty for exposure information missing from the representation. Since $\Sigma_f\succ0$, this penalty is zero if and only if $\delta_B=0$ almost surely, meaning that $B$ can be recovered from $z$. The conditional-mean penalty is zero if and only if $\delta_{\bar u}=0$ almost surely.

This decomposition holds for any fixed $z$. Thus, the minimum equals the floor $\E\|\varepsilon\|^2$ if and only if both $\bar u(x)$ and $B$ can be recovered from $z$. Since $m$ is fixed, recovering $\bar u(x)=\alpha+Bm$ and $B$ is equivalent to recovering $\alpha$ and $B$, completing the proof.

%% file: sections/B_appendix.tex
\section{Synthetic Panel Details}
\label{app:synthetic}
We draw $a_i\sim\mathcal{N}(0,0.5^2)$, $B_i\sim\mathcal{N}(1,0.5^2)$ and $e_{i,t}\sim\mathcal{N}(0,1)$. Factor predictability is set through an observed signal $s_t\sim\mathcal{N}(0,1)$, with $f_{t+1}=\rho s_t+\sqrt{1-\rho^2}\,\eta_{t+1}$ and $\eta$ i.i.d.\ standard normal. The past then predicts $f_{t+1}$ with $R^2=\rho^2$, while $f$ stays i.i.d.\ with unit variance, so identifying $B_i$ from history is equally hard at every $\rho$. The encoder is a two-layer MLP that sees the last 32 returns of the asset together with the factor and signal paths. The target is the next return $r_{i,t+1}$, so~\eqref{eq:factor} holds exactly, with an observed target in place of an EMA embedding. Per-asset JEPA uses $\lambda=0$. MC-JEPA uses $K=1$, $\lambda=0.75$, and the leave-one-out panel mean of the targets as the condition~\eqref{eq:condition}. Both train on 1,024 assets over 4,000 time steps, with 32 dates of 64 assets per step for 4,000 steps. For evaluation we simulate 1,024 new assets on a new factor path, embed 16 windows per asset with the frozen encoder, fit a probe on half of the assets and report $R^2$ on the other half, averaged over five seeds.

%% file: sections/C_appendix.tex
\section{Model and Training Hyperparameters}
\label{app:hparams}

Tables~\ref{tab:hp-arch} and~\ref{tab:hp-train} give the configuration of MC-JEPA. Every model we train, including the self-supervised baselines, uses the same Transformer backbone (Table~\ref{tab:hp-arch}, about 10M parameters), except that next-patch AR uses causal attention and no \texttt{[CLS]} token. All share the same 960-bar context and the optimizer and schedule of Table~\ref{tab:hp-train}, and are trained with data parallelism on two NVIDIA H800 GPUs.

\begin{table}[ht]
\centering
\caption{Encoder and predictor config.}
\label{tab:hp-arch}
\small
\begin{tabular}{ll}
\toprule
Setting & Value \\
\midrule
Patch length $p$ & 10 bars \\
Transformer layers & 12 \\
Width $D$ / MLP width & 256 / 1024 \\
Attention heads & 8 \\
Encoder parameters & 9.52M \\
\midrule
Horizons $\mathcal{H}$ & $\{1,60,240\}$ bars \\
Target block length $\Delta$ & 30 bars \\
Predictor MLP blocks ($a_\phi$, $m_\psi$) & 2 \\
Number of factors $K$ & 1 \\
\bottomrule
\end{tabular}
\end{table}

\begin{table}[ht]
\centering
\caption{Training settings of MC-JEPA.}
\label{tab:hp-train}
\small
\begin{tabular}{ll}
\toprule
Setting & Value \\
\midrule
Decision times per step & 128 \\
Condition pool $|\mathcal{M}_t|$ & 32 \\
Scored assets $|\mathcal{S}_t|$ & 8 \\
Condition weight $\lambda$ & 0.75 \\
\midrule
Optimizer & AdamW, $\beta=(0.9,0.95)$ \\
Learning rate & $3\times10^{-4}$ \\
Schedule & 1{,}000-step linear warm-up, cosine decay to 0 \\
Weight decay & 0.05 \\
Gradient clipping (norm) & 1.0 \\
Training steps & 20{,}000 \\
EMA momentum $\tau$ & linear from $\tau_0=0.996$ to 1 \\
Precision & bfloat16 \\
\bottomrule
\end{tabular}
\end{table}

\textbf{Self-supervised baselines.} Per-asset JEPA trains on 1024 independent windows per step, matching the 1024 scored windows of MC-JEPA. TS2Vec uses its official loss and augmentations, and CPC uses InfoNCE with temperature 0.1 over the per-GPU batch. JEPA variants and CPC are read out from the final \texttt{[CLS]} state, whose objective trains it, and all other baselines from the mean of their final patch states. For every baseline this is the better of the two readouts.

\textbf{Pretrained foundation models.} We evaluate each foundation model with frozen weights using its official implementation and native inference preprocessing. We extract the final-layer hidden states used by the model's forecasting or reconstruction head. Kronos takes full OHLCV bars, which it normalizes and tokenizes, and we average its hidden states over all bars. For TimesFM-2.5, Chronos-2 and MOMENT, we provide the close price and volume as two series, relying on each model's internal normalization. TimesFM-2.5 and MOMENT process the two series independently, whereas Chronos-2 treats them as a multivariate input, allowing cross-series interactions through group attention. For all three models, we average the final states over the window's patches for each series and concatenate the two series' embeddings. Each model reads the most recent bars of the window up to its maximum context length, capped at the 960-bar window. Table~\ref{tab:hp-tsfm} lists the checkpoints.

\begin{table}[ht]
\centering
\caption{Checkpoints of the pretrained time-series foundation model baselines.}
\label{tab:hp-tsfm}
\small
\begin{tabular}{ll}
\toprule
Model & Checkpoint \\
\midrule
Kronos-mini & \texttt{NeoQuasar/Kronos-mini} \\
Kronos-small & \texttt{NeoQuasar/Kronos-small} \\
Kronos-base & \texttt{NeoQuasar/Kronos-base} \\
TimesFM-2.5 & \texttt{google/timesfm-2.5-200m-pytorch} \\
Chronos-2 & \texttt{amazon/chronos-2} \\
MOMENT-small & \texttt{AutonLab/MOMENT-1-small} \\
MOMENT-base & \texttt{AutonLab/MOMENT-1-base} \\
\bottomrule
\end{tabular}
\end{table}

%% file: sections/D_appendix.tex
\section{Evaluation Targets and Metrics}
\label{app:metrics}

\subsection{Downstream Targets}
\label{app:metrics:targets}

One sample is one stock on one trading day $t$, and all targets below are defined for that stock. Let $P_{d}$ be its close price on day $d$, $V_{d}$ its total volume on day $d$, and $R_{d}=P_{d}/P_{d-1}-1$ its daily return. Daily targets are decided at the 15:00 close of day $t$, so the encoder window ends at that close. Minute targets are decided at the 11:30 morning close of day $t$. For them, let $p_{0}$ be the 11:30 close, $p_{k}$ the close of the $k$-th 1-minute bar after it, $r_{k}=p_{k}/p_{k-1}-1$, and $v_{k}$ the volume of that bar. The 120 bars after 11:30 form the whole afternoon session, and we treat the lunch break as one ordinary minute. A target is dropped if the stock does not trade on every market day it spans.

\begin{itemize}
\item \textbf{Ret $h$d} ($h\in\{1,5,20\}$): $P_{t+1+h}/P_{t+1}-1$. The return starts at the close of day $t+1$, so a signal formed at the close of $t$ can be traded at the next close, as in the backtest.
\item \textbf{Ret $k$m} ($k\in\{5,30,60,120\}$): $p_{k}/p_{1}-1$. The return starts after the first bar, so the bid-ask bounce and the lunch-break news in the 11:30 and 13:01 bars cannot drive it. Ret 120m is the afternoon return.
\item \textbf{Beta 20d}: the OLS slope of $R_{t+k}$ on $M_{t+k}$ over $k=1,\dots,20$, where $M_d$ is the equal-weight mean of the daily returns of all stocks on day $d$.
\item \textbf{Vol $k$m} ($k\in\{5,30,60,120\}$): the standard deviation of $r_{1},\dots,r_{k}$.
\item \textbf{Vol 1d}: the standard deviation of the 1-minute close-to-close returns of day $t+1$. The first of them includes the overnight gap.
\item \textbf{Vol 5d}: the same standard deviation, pooled over days $t+1,\dots,t+5$.
\item \textbf{MDD 5d}: the largest drop of the 1-minute close from its running peak over days $t+1,\dots,t+5$, with the peak starting at $P_t$.
\item \textbf{Volume $k$m} ($k\in\{5,30,60,120\}$): the mean of $v_{1},\dots,v_{k}$, divided by the mean volume per bar of day $t$ up to 11:30.
\item \textbf{Volume 1d}: $V_{t+1}/V_{t}$.
\item \textbf{Volume 5d}: $\frac{1}{5}\sum_{k=1}^{5}V_{t+k}/V_{t}$.
\item \textbf{Open vol share}: the share of $V_{t+1}$ traded in the first 30 bars of day $t+1$.
\end{itemize}

\subsection{RankIC}
\label{app:metrics:rankic}

Let $\mathcal{T}$ be the test days and $\mathcal{U}_d$ the stocks on day $d$ with both a prediction $\hat y_{i,d}$ and a realized target $y_{i,d}$. The daily RankIC is the Pearson correlation of the within-day ranks,
\begin{equation}
\rho_d=\operatorname{corr}_{i\in\mathcal{U}_d}\bigl(\operatorname{rank}_d(\hat y_{i,d}),\,\operatorname{rank}_d(y_{i,d})\bigr),
\end{equation}
with ties given their average rank. We report $\mathrm{RankIC}=\frac{1}{|\mathcal{T}'|}\sum_{d\in\mathcal{T}'}\rho_d$, where $\mathcal{T}'\subseteq\mathcal{T}$ keeps the days with $|\mathcal{U}_d|\ge20$.

\subsection{Backtest Metrics}
\label{app:metrics:backtest}

The backtest of Table~\ref{tab:backtest} trades the Ret 20d probe. On each signal day $t$, a new tranche buys the top 7 or 8 predictions among the point-in-time CSI300 members at the close of $t+1$ and sells them at the close of $t+21$. Each tranche receives $1/20$ of the capital with equal weights inside it, so the book holds 20 overlapping tranches of about 150 stocks in total. Weights are reset daily, and only the net weight change of each stock is traded, at 5\,bp on buys and 15\,bp on sells. Prices are daily back-adjusted closes from BaoStock. Suspended days are removed from the data, so a held stock earns zero return while suspended. Let $r_d$ be the daily return of the book net of costs and $b_d$ that of the CSI300 index, over the $N$ days from the first day the book holds all 20 tranches to the end of the test period. Following Qlib~\citep{Yang2020QlibAA}, we use arithmetic annualization with 238 trading days per year:
\begin{align}
\mathrm{AER}&=238\cdot\frac{1}{N}\sum_{d}(r_d-b_d),\\
\mathrm{Sharpe}&=\sqrt{238}\,\frac{\operatorname{mean}_d(r_d)}{\operatorname{std}_d(r_d)},\\
\mathrm{IR}&=\sqrt{238}\,\frac{\operatorname{mean}_d(r_d-b_d)}{\operatorname{std}_d(r_d-b_d)},\\
\mathrm{MDD}&=\max_{s\le d}\Bigl(-\sum_{u=s+1}^{d}r_u\Bigr).
\end{align}
AER and IR are measured in excess of the index, while Sharpe and MDD are computed on the net return of the book. The Sharpe ratio uses no risk-free rate, and the drawdown is taken on cumulative returns.

\begin{figure}[t]
\centering
\includegraphics[width=\linewidth]{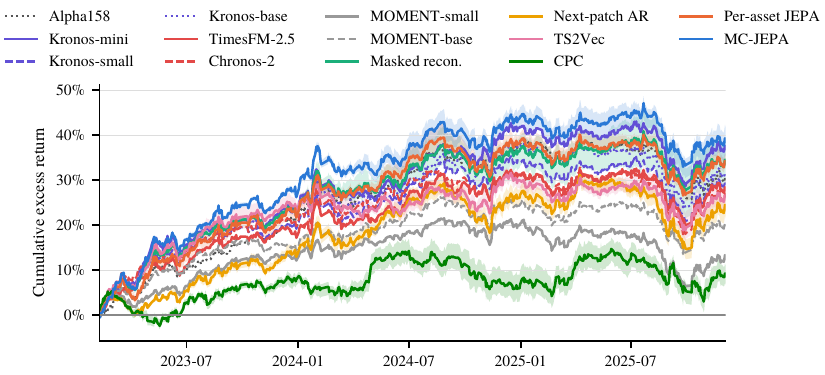}
\caption{Cumulative excess return over the CSI300 index of the backtest in Table~\ref{tab:backtest}, net of transaction costs. For the encoders trained in our experiments, lines show the mean and shaded bands the range over three seeds.}
\label{fig:backtest-curve}
\end{figure}

Figure~\ref{fig:backtest-curve} plots the cumulative excess return of every method in Table~\ref{tab:backtest} over the test period. MC-JEPA leads from early 2024 and keeps its lead through the 2025 drawdown, with Kronos-mini the closest.

%% file: sections/E_appendix.tex
\section{Additional Experiment Results}
\label{app:results}

Table~\ref{tab:minute-horizons} repeats the minute targets of Table~\ref{tab:main} at shorter horizons, and Table~\ref{tab:daily-extra} adds four daily targets. Appendix~\ref{app:metrics:targets} defines all of them. Probes, splits and seeds are the same as in Table~\ref{tab:main}.

\textbf{Minute horizons.} The gains of MC-JEPA on the minute targets hold at every horizon. It is the best encoder on volume and second to CPC on volatility at 5, 30 and 60 minutes, as it is at 120 minutes. On returns it is best at 5 and 60 minutes, and at 30 minutes it trails per-asset JEPA within one seed standard deviation.

\textbf{Additional daily targets.} The daily targets give a more mixed picture. MC-JEPA is the best encoder on Volume 5d, consistent with Volume 1d. The daily risk targets favor CPC, which is the best encoder on all three and the best method on Vol 1d and Vol 5d. MC-JEPA ranks fourth among encoders on MDD 5d, behind CPC, next-patch AR and Chronos-2, and sixth on Vol 1d and Vol 5d, where Kronos-base and TimesFM-2.5 also lead it. Compared with per-asset JEPA, conditioning improves every daily target.

\textbf{Ablations.} Tables~\ref{tab:ablation-minute} and~\ref{tab:ablation-daily} report the same targets for the variants of Table~\ref{tab:ablation}. Every variant with an aligned condition beats both per-asset and panel JEPA on all volatility, drawdown and volume targets, at every horizon. The placebo stays below MC-JEPA on every target except Ret 60m, where the two tie. More factors help the risk and liquidity targets but hurt daily returns. Every risk and liquidity target is best at some $K>1$. The number of factors trades risk and liquidity prediction against return prediction. Pool size matters most at minute horizons. Vol 5m, Vol 30m and Vol 60m rise with $|\mathcal{M}_t|$, as Vol 120m does, and $|\mathcal{M}_t|=128$ is above MC-JEPA on all three minute return horizons, although its Ret 30m varies widely across seeds. The daily targets change little with pool size.

\textbf{Hyperparameter selection.} Table~\ref{tab:ablation-val} reports the condition weight, factor-count and condition pool variants of Table~\ref{tab:ablation} on the validation period, 2022-07 to 2022-12. We select $\lambda$ and $K$ by Ret 1d and Ret 5d on this period. $\lambda=0.75$ and $K=1$ are best on both. Larger $K$ improves Beta 20d and Vol 120m, and $\lambda\ge0.75$ matches or improves on smaller weights on every target except Ret 120m. Among pool sizes, $|\mathcal{M}_t|=32$ is best or tied on seven of the eight targets.

\textbf{Statistical significance.} Table~\ref{tab:paired} tests whether the gains of MC-JEPA over per-asset JEPA and over the placebo condition are robust to day-to-day variation, and reports their consistency across seeds. For each target, we average the daily test RankIC of each method over its three seeds and compute the two-sided Newey--West $t$-statistic of the mean daily difference. Overlapping labels make the daily RankIC autocorrelated, so a target whose label spans $h$ trading days uses $\max(h,6)$ lags, where 6 is the automatic choice $\lfloor 4(T/100)^{2/9}\rfloor$ for $T\approx720$ test days. This gives 20 lags for Ret 20d and Beta 20d and 6 for every other target. We also count the seeds for which the MC-JEPA run beats the run of the other method with the same seed. We report unadjusted tests at the 5\% level.

Against both methods, MC-JEPA is significantly better on every exposure, risk and liquidity target, with all three seed pairs in its favor. Return gains are smaller and less uniform. On daily returns, the gains over the placebo are significant on Ret 1d, Ret 5d and Ret 20d, and the gains over per-asset JEPA are significant on Ret 1d and Ret 20d but not on Ret 5d ($t=1.8$), with all seed pairs in favor of MC-JEPA in each case. Ret 20d is the closest to the threshold, with $t=2.2$ and $2.0$. Minute returns show no reliable gain. None of the four horizons is significant against per-asset JEPA, and against the placebo only Ret 120m is.

\begin{figure}[t]
\centering
\includegraphics[width=\linewidth]{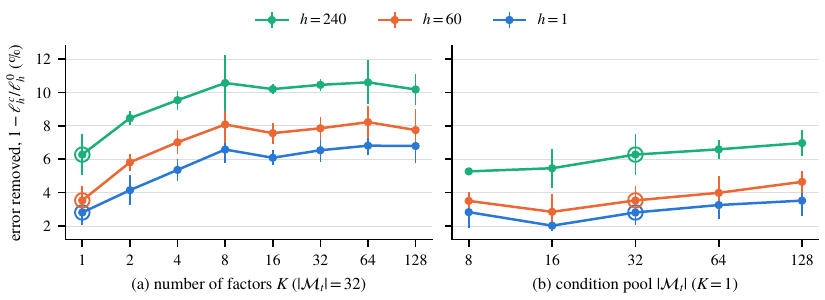}
\caption{Share of the unconditional error removed by the market condition at each horizon $h$, $1-\ell^{c}_h/\ell^{0}_h$, on the validation period at the end of training, (a) by number of factors $K$ at $|\mathcal{M}_t|=32$ and (b) by condition pool size $|\mathcal{M}_t|$ at $K=1$. Points show the mean and error bars the standard deviation over three seeds. The circled point is MC-JEPA in both panels.}
\label{fig:condition-reduction}
\end{figure}

\phantomsection\label{app:denoise}
\textbf{Measured reduction.} Under~\eqref{eq:factor}, the error of the optimal unconditional predictor is $B_i(f_h-m)+\varepsilon_{i,h}$, whereas the error of the optimal conditional predictor is $\varepsilon_{i,h}$ alone. Therefore, the conditional mean $\bar u(x_i)$ is learned against noise of variance $\E\|\varepsilon_{i,h}\|^2$ instead of $\E\|\varepsilon_{i,h}\|^2+\E\|B_i(f_h-m)\|^2$. We measure the share of the unconditional error that the learned condition removes, $1-\ell^{c}_h/\ell^{0}_h$, where $\ell^{0}_h$ and $\ell^{c}_h$ are the two terms of~\eqref{eq:objective} at horizon $h$. On the 2022-07 to 2022-12 validation period, at the end of training, the panel mean of the target embeddings accounts for $26\%$ of their variance, yet the condition of MC-JEPA ($K=1$) removes only $4.5\pm0.9\%$ of the unconditional error over all horizons. Figure~\ref{fig:condition-reduction} shows how this share changes with the horizon, the number of factors and the condition pool. It is largest at the longest horizon, where market moves account for more of an asset's future. It rises with $K$ up to $K=8$ and then levels off below $9\%$, and it grows with the pool mainly at the longer horizons, consistent with a less noisy market estimate. Therefore, in the target-embedding space the realized market explains only a small part of the error that remains after conditioning on the asset's own history, even with more factors, so the denoising effect on real data is modest. Return prediction does not improve with the share of error removed, as $K=1$ removes the least error yet achieves the best return prediction performance. As in the synthetic study, denoising does not account for all of the return gain. In the synthetic panel, a predictor that knows $a_i$, $B_i$ and $\E[f_{t+1}\mid\text{past}]$ is left with unconditional error $\E[B_i^2](1-R^2)+\operatorname{var}(e)$, where $R^2$ is the predictability of $f_{t+1}$ from the past, and observing $f_{t+1}$ removes the first term. The best achievable reduction falls from $56\%$ to $11\%$ as $R^2$ rises from 0 to 0.9 ($\E[B_i^2]=1.25$, $\operatorname{var}(e)=1$). The learned condition removes $30\pm4\%$ and $5\pm4\%$ at these two ends, about half of the best achievable reduction.

\begin{table}[t]
\centering
\caption{RankIC of the minute targets at horizons of 5, 30 and 60 minutes, decided at the 11:30 morning close. The 120-minute horizon is in Table~\ref{tab:main}. Encoders trained in our experiments report mean$_{\pm\mathrm{std}}$ over three seeds. \textbf{Bold} and \underline{underlined} values indicate the best and second-best scores, respectively.}
\label{tab:minute-horizons}
\newcommand{\s}[1]{$_{\pm#1}$}
\setlength{\tabcolsep}{3pt}
\resizebox{\textwidth}{!}{%
\begin{tabular}{l ccc ccc ccc}
\toprule
 & \multicolumn{3}{c}{Return} & \multicolumn{3}{c}{Risk} & \multicolumn{3}{c}{Liquidity} \\
\cmidrule(lr){2-4}\cmidrule(lr){5-7}\cmidrule(lr){8-10}
Method & Ret 5m & Ret 30m & Ret 60m & Vol 5m & Vol 30m & Vol 60m & Volume 5m & Volume 30m & Volume 60m \\
\midrule
Alpha158 & .0459 & .0629 & .0607 & .256 & .347 & .359 & .172 & .195 & .223 \\
\midrule
Kronos-mini & .0546 & .0682 & .0625 & .146 & .339 & .371 & .292 & .358 & .407 \\
Kronos-small & .0600 & .0683 & .0637 & .188 & .364 & .395 & .366 & .441 & .489 \\
Kronos-base & .0603 & .0686 & .0634 & .207 & .394 & .425 & .381 & .455 & .502 \\
TimesFM-2.5 & .0599 & .0727 & .0696 & .159 & .352 & .387 & .343 & .410 & .457 \\
Chronos-2 & .0719 & .0717 & .0677 & .252 & .442 & .469 & \underline{.430} & \underline{.487} & \underline{.525} \\
MOMENT-small & .0628 & .0701 & .0685 & .135 & .285 & .313 & .362 & .429 & .470 \\
MOMENT-base & .0635 & .0683 & .0694 & .139 & .282 & .307 & .357 & .426 & .468 \\
\midrule
Masked recon. & .0631\s{.0014} & .0749\s{.0004} & .0695\s{.0011} & .136\s{.011} & .273\s{.013} & .296\s{.011} & .314\s{.017} & .381\s{.018} & .429\s{.017} \\
Next-patch AR & .0645\s{.0016} & .0748\s{.0017} & .0671\s{.0021} & .179\s{.005} & .343\s{.006} & .371\s{.007} & .350\s{.006} & .419\s{.004} & .470\s{.002} \\
TS2Vec & .0630\s{.0006} & .0742\s{.0018} & \underline{.0704}\s{.0009} & .146\s{.008} & .323\s{.013} & .356\s{.014} & .303\s{.005} & .368\s{.005} & .416\s{.005} \\
CPC & .0720\s{.0010} & .0737\s{.0010} & .0674\s{.0017} & \textbf{.293}\s{.002} & \textbf{.502}\s{.004} & \textbf{.531}\s{.005} & .419\s{.001} & .463\s{.002} & .496\s{.002} \\
Per-asset JEPA & \underline{.0730}\s{.0025} & \textbf{.0767}\s{.0036} & \textbf{.0716}\s{.0054} & .240\s{.011} & .408\s{.017} & .430\s{.017} & .424\s{.002} & .481\s{.003} & .517\s{.004} \\
\midrule
MC-JEPA & \textbf{.0742}\s{.0014} & \underline{.0754}\s{.0020} & \textbf{.0716}\s{.0022} & \underline{.275}\s{.004} & \underline{.476}\s{.006} & \underline{.500}\s{.006} & \textbf{.441}\s{.003} & \textbf{.500}\s{.005} & \textbf{.538}\s{.005} \\
\bottomrule
\end{tabular}}
\end{table}

\begin{table}[t]
\centering
\caption{RankIC of additional daily targets, decided at the 15:00 close. Encoders trained in our experiments report mean$_{\pm\mathrm{std}}$ over three seeds. \textbf{Bold} and \underline{underlined} values indicate the best and second-best scores, respectively.}
\label{tab:daily-extra}
\newcommand{\s}[1]{$_{\pm#1}$}
\setlength{\tabcolsep}{3pt}
\begin{tabular}{l ccc c}
\toprule
 & \multicolumn{3}{c}{Risk} & Liquidity \\
\cmidrule(lr){2-4}\cmidrule(lr){5-5}
Method & Vol 1d & Vol 5d & MDD 5d & Volume 5d \\
\midrule
Alpha158 & \underline{.582} & \underline{.562} & \textbf{.590} & .495 \\
\midrule
Kronos-mini & .540 & .522 & .492 & .451 \\
Kronos-small & .540 & .518 & .505 & .496 \\
Kronos-base & .551 & .532 & .507 & .504 \\
TimesFM-2.5 & .551 & .537 & .520 & .486 \\
Chronos-2 & .575 & .557 & .527 & .511 \\
MOMENT-small & .477 & .452 & .491 & .457 \\
MOMENT-base & .468 & .446 & .487 & .454 \\
\midrule
Masked recon. & .465\s{.028} & .452\s{.030} & .500\s{.008} & .473\s{.003} \\
Next-patch AR & .557\s{.005} & .543\s{.006} & .529\s{.005} & .499\s{.001} \\
TS2Vec & .527\s{.004} & .512\s{.004} & .509\s{.002} & .456\s{.005} \\
CPC & \textbf{.601}\s{.003} & \textbf{.581}\s{.002} & \underline{.545}\s{.002} & .489\s{.004} \\
Per-asset JEPA & .512\s{.007} & .495\s{.007} & .516\s{.000} & \underline{.513}\s{.003} \\
\midrule
MC-JEPA & .543\s{.003} & .525\s{.004} & .525\s{.002} & \textbf{.528}\s{.005} \\
\bottomrule
\end{tabular}
\end{table}

\begin{table}[t]
\centering
\caption{Ablation results on the minute targets of Table~\ref{tab:minute-horizons}. Variants and seeds are as in Table~\ref{tab:ablation}. \textbf{Bold} values indicate the best score in each column.}
\label{tab:ablation-minute}
\newcommand{\s}[1]{$_{\pm#1}$}
\setlength{\tabcolsep}{3pt}
\resizebox{\textwidth}{!}{%
\begin{tabular}{l ccc ccc ccc}
\toprule
 & \multicolumn{3}{c}{Return} & \multicolumn{3}{c}{Risk} & \multicolumn{3}{c}{Liquidity} \\
\cmidrule(lr){2-4}\cmidrule(lr){5-7}\cmidrule(lr){8-10}
Variant & Ret 5m & Ret 30m & Ret 60m & Vol 5m & Vol 30m & Vol 60m & Volume 5m & Volume 30m & Volume 60m \\
\midrule
\multicolumn{10}{l}{\emph{(a) Components}} \\
Per-asset JEPA & .0730\s{.0025} & .0767\s{.0036} & .0716\s{.0054} & .240\s{.011} & .408\s{.017} & .430\s{.017} & .424\s{.002} & .481\s{.003} & .517\s{.004} \\
Panel JEPA ($\lambda=0$) & .0718\s{.0008} & .0758\s{.0024} & .0717\s{.0036} & .243\s{.006} & .408\s{.023} & .426\s{.027} & .425\s{.003} & .483\s{.004} & .518\s{.004} \\
Placebo condition (other date) & .0729\s{.0017} & .0738\s{.0025} & .0716\s{.0065} & .262\s{.010} & .447\s{.019} & .470\s{.021} & .432\s{.005} & .491\s{.004} & .528\s{.003} \\
Unfactored predictor & .0707\s{.0017} & .0730\s{.0013} & .0695\s{.0058} & .270\s{.004} & .462\s{.012} & .483\s{.013} & .437\s{.004} & .497\s{.005} & .534\s{.005} \\
\midrule
\multicolumn{10}{l}{\emph{(b) Condition weight $\lambda$}} \\
$\lambda=0.25$ & .0729\s{.0011} & .0743\s{.0019} & .0695\s{.0009} & .269\s{.004} & .464\s{.003} & .487\s{.003} & .438\s{.003} & .496\s{.001} & .533\s{.001} \\
$\lambda=0.5$ & .0739\s{.0022} & .0763\s{.0012} & .0695\s{.0007} & .280\s{.006} & .478\s{.009} & .502\s{.011} & .438\s{.004} & .497\s{.006} & .533\s{.006} \\
$\lambda=1$ & \textbf{.0757}\s{.0018} & .0758\s{.0016} & \textbf{.0741}\s{.0044} & .284\s{.007} & .481\s{.010} & .505\s{.009} & .444\s{.003} & .502\s{.006} & .538\s{.006} \\
\midrule
\multicolumn{10}{l}{\emph{(c) Number of factors $K$}} \\
$K=2$ & .0737\s{.0022} & .0749\s{.0009} & .0683\s{.0019} & .285\s{.002} & .485\s{.007} & .509\s{.009} & .446\s{.003} & .504\s{.004} & .541\s{.004} \\
$K=4$ & .0735\s{.0007} & .0754\s{.0021} & .0691\s{.0006} & \textbf{.295}\s{.005} & \textbf{.497}\s{.014} & \textbf{.522}\s{.016} & .447\s{.004} & .506\s{.003} & .543\s{.003} \\
$K=8$ & .0729\s{.0010} & .0741\s{.0022} & .0678\s{.0010} & .282\s{.009} & .481\s{.007} & .506\s{.006} & .445\s{.001} & .504\s{.003} & .542\s{.003} \\
$K=16$ & .0745\s{.0020} & \textbf{.0770}\s{.0016} & .0735\s{.0016} & .291\s{.003} & .495\s{.003} & .520\s{.003} & .445\s{.001} & .505\s{.001} & .542\s{.001} \\
$K=32$ & .0745\s{.0013} & .0764\s{.0015} & .0699\s{.0011} & .290\s{.003} & \textbf{.497}\s{.002} & \textbf{.522}\s{.002} & .446\s{.002} & .505\s{.001} & .543\s{.001} \\
$K=64$ & .0734\s{.0006} & .0756\s{.0017} & .0703\s{.0022} & .288\s{.010} & .493\s{.010} & .518\s{.011} & \textbf{.448}\s{.000} & \textbf{.507}\s{.003} & \textbf{.545}\s{.004} \\
$K=128$ & .0739\s{.0024} & .0760\s{.0027} & \textbf{.0741}\s{.0017} & .287\s{.008} & .495\s{.012} & .519\s{.012} & \textbf{.448}\s{.002} & \textbf{.507}\s{.002} & \textbf{.545}\s{.003} \\
\midrule
\multicolumn{10}{l}{\emph{(d) Condition pool size $|\mathcal{M}_t|$}} \\
$|\mathcal{M}_t|=8$ & .0735\s{.0017} & .0759\s{.0000} & .0711\s{.0039} & .271\s{.006} & .465\s{.007} & .490\s{.008} & .443\s{.002} & .500\s{.001} & .537\s{.000} \\
$|\mathcal{M}_t|=16$ & .0720\s{.0030} & .0736\s{.0009} & .0687\s{.0013} & .272\s{.012} & .468\s{.018} & .491\s{.020} & .441\s{.002} & .499\s{.003} & .536\s{.003} \\
$|\mathcal{M}_t|=64$ & .0723\s{.0018} & .0757\s{.0002} & .0710\s{.0025} & .277\s{.004} & .475\s{.008} & .499\s{.008} & .440\s{.003} & .498\s{.003} & .535\s{.003} \\
$|\mathcal{M}_t|=128$ & .0753\s{.0017} & .0763\s{.0037} & .0725\s{.0007} & .281\s{.004} & .480\s{.002} & .505\s{.001} & .442\s{.003} & .501\s{.004} & .537\s{.005} \\
\midrule
\begin{tabular}[t]{@{}l@{}}MC-JEPA\\($\lambda=0.75$, $K=1$, $|\mathcal{M}_t|=32$)\end{tabular} & .0742\s{.0014} & .0754\s{.0020} & .0716\s{.0022} & .275\s{.004} & .476\s{.006} & .500\s{.006} & .441\s{.003} & .500\s{.005} & .538\s{.005} \\
\bottomrule
\end{tabular}}
\end{table}

\begin{table}[t]
\centering
\caption{Ablation results on the daily targets of Table~\ref{tab:daily-extra}. Variants and seeds are as in Table~\ref{tab:ablation}. \textbf{Bold} values indicate the best score in each column.}
\label{tab:ablation-daily}
\newcommand{\s}[1]{$_{\pm#1}$}
\setlength{\tabcolsep}{3pt}
\begin{tabular}{l ccc c}
\toprule
 & \multicolumn{3}{c}{Risk} & Liquidity \\
\cmidrule(lr){2-4}\cmidrule(lr){5-5}
Variant & Vol 1d & Vol 5d & MDD 5d & Volume 5d \\
\midrule
\multicolumn{5}{l}{\emph{(a) Components}} \\
Per-asset JEPA & .512\s{.007} & .495\s{.007} & .516\s{.000} & .513\s{.003} \\
Panel JEPA ($\lambda=0$) & .485\s{.015} & .467\s{.016} & .510\s{.004} & .513\s{.002} \\
Placebo condition (other date) & .517\s{.012} & .499\s{.011} & .518\s{.003} & .522\s{.005} \\
Unfactored predictor & .527\s{.004} & .507\s{.004} & .520\s{.001} & .524\s{.005} \\
\midrule
\multicolumn{5}{l}{\emph{(b) Condition weight $\lambda$}} \\
$\lambda=0.25$ & .537\s{.011} & .518\s{.012} & .524\s{.001} & .523\s{.003} \\
$\lambda=0.5$ & .537\s{.009} & .515\s{.008} & .522\s{.002} & .524\s{.003} \\
$\lambda=1$ & .547\s{.005} & .527\s{.007} & .524\s{.001} & .527\s{.001} \\
\midrule
\multicolumn{5}{l}{\emph{(c) Number of factors $K$}} \\
$K=2$ & .557\s{.012} & .537\s{.014} & .527\s{.003} & .528\s{.005} \\
$K=4$ & .564\s{.006} & \textbf{.546}\s{.007} & .528\s{.004} & .530\s{.003} \\
$K=8$ & .554\s{.008} & .534\s{.007} & .528\s{.002} & .531\s{.004} \\
$K=16$ & .559\s{.005} & .540\s{.005} & .529\s{.001} & .530\s{.001} \\
$K=32$ & \textbf{.565}\s{.005} & \textbf{.546}\s{.005} & \textbf{.530}\s{.002} & .530\s{.002} \\
$K=64$ & .562\s{.004} & .541\s{.004} & .529\s{.002} & \textbf{.532}\s{.002} \\
$K=128$ & .558\s{.005} & .538\s{.005} & .526\s{.001} & \textbf{.532}\s{.001} \\
\midrule
\multicolumn{5}{l}{\emph{(d) Condition pool size $|\mathcal{M}_t|$}} \\
$|\mathcal{M}_t|=8$ & .542\s{.008} & .522\s{.007} & .522\s{.001} & .526\s{.002} \\
$|\mathcal{M}_t|=16$ & .537\s{.008} & .517\s{.008} & .522\s{.002} & .525\s{.001} \\
$|\mathcal{M}_t|=64$ & .547\s{.009} & .528\s{.009} & .524\s{.004} & .526\s{.004} \\
$|\mathcal{M}_t|=128$ & .542\s{.010} & .523\s{.011} & .525\s{.004} & .526\s{.003} \\
\midrule
\begin{tabular}[t]{@{}l@{}}MC-JEPA\\($\lambda=0.75$, $K=1$, $|\mathcal{M}_t|=32$)\end{tabular} & .543\s{.003} & .525\s{.004} & .525\s{.002} & .528\s{.005} \\
\bottomrule
\end{tabular}
\end{table}

\begin{table}[t]
\centering
\caption{Validation results (2022-07 to 2022-12) for the condition weight, factor-count and condition pool variants of Table~\ref{tab:ablation}, on its targets. Probes are fitted on 2019-01 to 2022-06, as in Table~\ref{tab:ablation}. Seeds are as in Table~\ref{tab:ablation}. \textbf{Bold} values indicate the best score in each column.}
\label{tab:ablation-val}
\newcommand{\s}[1]{$_{\pm#1}$}
\setlength{\tabcolsep}{3pt}
\resizebox{\textwidth}{!}{%
\begin{tabular}{l ccc c c ccc}
\toprule
 & \multicolumn{3}{c}{Return} & Exposure & Risk & \multicolumn{3}{c}{Liquidity} \\
\cmidrule(lr){2-4}\cmidrule(lr){5-5}\cmidrule(lr){6-6}\cmidrule(lr){7-9}
Variant & Ret 120m & Ret 1d & Ret 5d & Beta 20d & Vol 120m & Volume 120m & Volume 1d & Open vol share \\
\midrule
\multicolumn{9}{l}{\emph{(a) Condition weight $\lambda$}} \\
$\lambda=0.25$ & .0620\s{.0043} & .0707\s{.0021} & .1146\s{.0013} & .311\s{.001} & .499\s{.002} & .614\s{.002} & .474\s{.003} & .405\s{.001} \\
$\lambda=0.5$ & .0671\s{.0031} & .0713\s{.0035} & .1159\s{.0051} & .312\s{.002} & .514\s{.008} & .614\s{.007} & .474\s{.002} & .405\s{.001} \\
$\lambda=1$ & .0649\s{.0032} & .0732\s{.0022} & .1171\s{.0023} & .319\s{.003} & .517\s{.009} & .619\s{.004} & .477\s{.001} & .406\s{.002} \\
\midrule
\multicolumn{9}{l}{\emph{(b) Number of factors $K$}} \\
$K=2$ & .0656\s{.0024} & .0700\s{.0025} & .1160\s{.0007} & .323\s{.002} & .521\s{.008} & .621\s{.003} & .477\s{.004} & .407\s{.002} \\
$K=4$ & \textbf{.0676}\s{.0031} & .0683\s{.0032} & .1167\s{.0028} & .323\s{.004} & .525\s{.017} & .624\s{.001} & .480\s{.001} & .407\s{.001} \\
$K=8$ & .0645\s{.0016} & .0710\s{.0039} & .1168\s{.0026} & .319\s{.002} & .519\s{.012} & .621\s{.003} & .480\s{.004} & .407\s{.002} \\
$K=16$ & .0652\s{.0026} & .0692\s{.0030} & .1161\s{.0026} & .323\s{.003} & .529\s{.007} & .623\s{.002} & .477\s{.002} & \textbf{.408}\s{.001} \\
$K=32$ & .0658\s{.0012} & .0681\s{.0029} & .1146\s{.0030} & \textbf{.324}\s{.002} & \textbf{.537}\s{.003} & .623\s{.001} & .479\s{.003} & \textbf{.408}\s{.001} \\
$K=64$ & .0649\s{.0016} & .0717\s{.0017} & .1172\s{.0015} & .321\s{.003} & .534\s{.010} & \textbf{.626}\s{.003} & \textbf{.482}\s{.001} & .407\s{.001} \\
$K=128$ & .0647\s{.0024} & .0679\s{.0031} & .1139\s{.0033} & .323\s{.002} & .531\s{.009} & .625\s{.002} & \textbf{.482}\s{.001} & \textbf{.408}\s{.001} \\
\midrule
\multicolumn{9}{l}{\emph{(c) Condition pool size $|\mathcal{M}_t|$}} \\
$|\mathcal{M}_t|=8$ & .0643\s{.0023} & .0712\s{.0019} & .1152\s{.0028} & .311\s{.003} & .495\s{.008} & .618\s{.001} & .476\s{.002} & .404\s{.001} \\
$|\mathcal{M}_t|=16$ & .0647\s{.0052} & \textbf{.0743}\s{.0024} & .1171\s{.0019} & .317\s{.004} & .497\s{.021} & .618\s{.003} & .476\s{.001} & .405\s{.001} \\
$|\mathcal{M}_t|=64$ & .0621\s{.0019} & .0700\s{.0004} & .1155\s{.0020} & .315\s{.002} & .511\s{.012} & .616\s{.004} & .477\s{.004} & .406\s{.001} \\
$|\mathcal{M}_t|=128$ & .0647\s{.0012} & .0711\s{.0027} & .1155\s{.0028} & .319\s{.001} & .515\s{.003} & .618\s{.005} & .477\s{.004} & .404\s{.001} \\
\midrule
\begin{tabular}[t]{@{}l@{}}MC-JEPA\\($\lambda=0.75$, $K=1$, $|\mathcal{M}_t|=32$)\end{tabular} & .0649\s{.0020} & .0735\s{.0010} & \textbf{.1182}\s{.0029} & .319\s{.002} & .516\s{.006} & .619\s{.007} & .478\s{.005} & .406\s{.001} \\
\bottomrule
\end{tabular}}
\end{table}

\begin{table}[t]
\centering
\caption{Paired tests of MC-JEPA against per-asset JEPA and the placebo condition on the 2023--2025 test set. $\Delta$ is the mean daily RankIC difference, and $t$ is its Newey--West $t$-statistic, with 20 lags for Ret 20d and Beta 20d and 6 otherwise. Bold marks $t>1.96$, a significant improvement at the 5\% level. Seeds is the number of seeds, out of three, for which MC-JEPA has the higher RankIC.}
\label{tab:paired}
\setlength{\tabcolsep}{4pt}
\begin{tabular}{l ccc ccc}
\toprule
 & \multicolumn{3}{c}{vs.\ Per-asset JEPA} & \multicolumn{3}{c}{vs.\ Placebo condition} \\
\cmidrule(lr){2-4}\cmidrule(lr){5-7}
Target & $\Delta$ & $t$ & Seeds & $\Delta$ & $t$ & Seeds \\
\midrule
\multicolumn{7}{l}{\emph{Return}} \\
Ret 5m & 0.0011 & 1.3 & 2/3 & 0.0013 & 1.8 & 2/3 \\
Ret 30m & $-$0.0013 & $-$1.1 & 1/3 & 0.0016 & 1.7 & 1/3 \\
Ret 60m & 0.0000 & 0.0 & 1/3 & 0.0000 & 0.0 & 1/3 \\
Ret 120m & 0.0018 & 1.5 & 2/3 & 0.0028 & \textbf{3.2} & 2/3 \\
Ret 1d & 0.0034 & \textbf{2.3} & 3/3 & 0.0026 & \textbf{5.2} & 3/3 \\
Ret 5d & 0.0036 & 1.8 & 3/3 & 0.0042 & \textbf{6.5} & 3/3 \\
Ret 20d & 0.0052 & \textbf{2.2} & 3/3 & 0.0039 & \textbf{2.0} & 3/3 \\
\midrule
\multicolumn{7}{l}{\emph{Exposure}} \\
Beta 20d & 0.0149 & \textbf{6.9} & 3/3 & 0.0125 & \textbf{8.6} & 3/3 \\
\midrule
\multicolumn{7}{l}{\emph{Risk}} \\
Vol 5m & 0.0345 & \textbf{31.8} & 3/3 & 0.0131 & \textbf{18.1} & 3/3 \\
Vol 30m & 0.0672 & \textbf{47.0} & 3/3 & 0.0281 & \textbf{30.2} & 3/3 \\
Vol 60m & 0.0700 & \textbf{48.6} & 3/3 & 0.0297 & \textbf{30.5} & 3/3 \\
Vol 120m & 0.0730 & \textbf{49.4} & 3/3 & 0.0303 & \textbf{30.3} & 3/3 \\
Vol 1d & 0.0310 & \textbf{28.4} & 3/3 & 0.0261 & \textbf{23.8} & 3/3 \\
Vol 5d & 0.0299 & \textbf{26.3} & 3/3 & 0.0260 & \textbf{21.4} & 3/3 \\
MDD 5d & 0.0083 & \textbf{10.7} & 3/3 & 0.0068 & \textbf{12.4} & 3/3 \\
\midrule
\multicolumn{7}{l}{\emph{Liquidity}} \\
Volume 5m & 0.0171 & \textbf{43.2} & 3/3 & 0.0093 & \textbf{30.2} & 3/3 \\
Volume 30m & 0.0198 & \textbf{42.1} & 3/3 & 0.0091 & \textbf{24.1} & 3/3 \\
Volume 60m & 0.0205 & \textbf{42.4} & 3/3 & 0.0099 & \textbf{23.2} & 3/3 \\
Volume 120m & 0.0210 & \textbf{39.8} & 3/3 & 0.0096 & \textbf{23.0} & 3/3 \\
Volume 1d & 0.0134 & \textbf{25.2} & 3/3 & 0.0059 & \textbf{15.1} & 3/3 \\
Volume 5d & 0.0153 & \textbf{24.3} & 3/3 & 0.0064 & \textbf{14.6} & 3/3 \\
Open vol share & 0.0064 & \textbf{18.1} & 3/3 & 0.0049 & \textbf{15.4} & 3/3 \\
\bottomrule
\end{tabular}
\end{table}

%% file: sections/F_appendix.tex
\section{Market Exposure in the Representation and Factor Head}
\label{app:exposure}

With $K=1$, the factor-structured predictor in~\eqref{eq:predictor} gives each stock a loading vector $b_\psi(z_i,h)\in\R^D$ on a single market factor. We examine whether these loadings encode market exposure and whether they can be interpreted as beta directly. All results use three MC-JEPA seeds and the 2023--2025 test period.

\textbf{Probed exposure.} Table~\ref{tab:exposure} reports ridge probes for Beta 20d on the representation $z_i$, on the loadings $b_\psi(z_i,240)$ at the longest horizon, and on their concatenation. The loadings nearly match the representation (0.366 versus 0.374), while combining them yields only a small improvement to 0.377. Since the loadings are a function of $z_i$, the factor head makes exposure information accessible to a linear probe rather than introducing new information.

Historical betas estimated over trailing 20, 60, and 120 days against the equal-weight market are substantially stronger (0.525). Combining them with $z_i$ improves RankIC to 0.539, consistently across seeds. Therefore, learned representation provides complementary exposure information.

\begin{table}[h]
\centering
\caption{Ridge probes for Beta 20d on the test period. Trailing betas are estimated against the equal-weight market and do not depend on the seed. Mean$_{\pm\mathrm{std}}$ over three seeds.}
\label{tab:exposure}
\small
\newcommand{\s}[1]{$_{\pm#1}$}
\begin{tabular}{lc}
\toprule
Features & Beta 20d \\
\midrule
\multicolumn{2}{l}{\emph{MC-JEPA}} \\
Representation $z_i$ & 0.374\s{0.003} \\
Loadings $b_\psi(z_i,240)$ & 0.366\s{0.003} \\
$z_i$ + loadings & 0.377\s{0.002} \\
\midrule
Trailing betas & 0.525 \\
MC-JEPA $z_i$ + trailing betas & 0.539\s{0.001} \\
\bottomrule
\end{tabular}
\end{table}

\textbf{The raw loadings measure co-movement.} We construct a label-free exposure estimate $\hat\beta_i = \langle b_i, \bar b\rangle / \lVert \bar b\rVert^2$, where $\bar b$ is the mean loading across stocks on the same date. The estimate varies substantially across stocks, with a cross-sectional standard deviation of 0.33--0.43, yet its RankIC with realized Beta 20d is only $-0.06$. To see what it ranks instead, we compare it with two statistics of each stock's daily returns over days $t{+}1$ to $t{+}20$. Its RankIC is 0.18 with the Pearson correlation between the stock and the equal-weight market, and $-0.26$ with the stock's volatility, the standard deviation of its returns. The correlation is not carried over from the context window: after regressing out the ranks of the window's return standard deviation and its trailing four-day correlation with the equal-weight market, computed from 5-minute returns, the partial RankIC is $0.105_{\pm0.008}$.

This discrepancy is consistent with the per-asset normalization in Section~\ref{sec:method:jepa}. Each asset's context and future bars are standardized using its own context-window statistics, reducing the information about absolute volatility scales. Since $\beta_i = \rho_i\,\sigma_i/\sigma_m$, where $\rho_i$ is the asset-market return correlation and $\sigma_i$ and $\sigma_m$ are the return volatilities of the asset and market respectively, capturing market co-movement $\rho_i$ alone does not suffice to recover beta. Therefore, the raw loadings should not be read as a beta directly.

\textbf{Restoring the removed scale.} The normalization standardizes each feature by the mean and standard deviation of its context window. We restore the removed scale by multiplying $\hat\beta_i$ by the coefficient of variation of the close price over the window, $\mathrm{CV}_i = \operatorname{std}(\text{close})/\operatorname{mean}(\text{close})$. RankIC is computed across stocks on each date, so the market volatility $\sigma_m$ is a per-date constant and does not need to be estimated. The result uses no labels, no fitted parameters and only the stock's own bars. Table~\ref{tab:exposure-scale} compares it with alternatives. With this scale, RankIC with Beta 20d rises from $-0.060$ to 0.368, close to the ridge probe on $z_i$ in Table~\ref{tab:exposure}, and 0.08 above $\mathrm{CV}_i$ alone. Therefore, the loadings add exposure information beyond the window's scale. The trailing four-day beta against the equal-weight market is stronger, but it requires the returns of all other stocks at inference.

\begin{table}[h]
\centering
\caption{Label-free beta estimates on the test period: daily RankIC with Beta 20d. $\hat\beta_i$ is the projection of the loading onto the date's mean loading. Window statistics are computed on the stock's own context window. The trailing four-day beta uses 5-minute returns of all stocks over the same window. Mean$_{\pm\mathrm{std}}$ over three seeds.}
\label{tab:exposure-scale}
\small
\newcommand{\s}[1]{$_{\pm#1}$}
\begin{tabular}{lc}
\toprule
Estimate & Beta 20d \\
\midrule
$\hat\beta_i$ & $-$0.060\s{0.011} \\
$\hat\beta_i\times\mathrm{CV}_i$ & 0.368\s{0.010} \\
\midrule
$\mathrm{CV}_i$ & 0.287 \\
Trailing 4-day beta & 0.434 \\
\bottomrule
\end{tabular}
\end{table}

%% file: references.bib
@article{assran2023self,
  title={Self-Supervised Learning from Images with a Joint-Embedding Predictive Architecture},
  author={Assran, Mahmoud and Duval, Quentin and Misra, Ishan and Bojanowski, Piotr and Vincent, Pascal and Rabbat, Michael and LeCun, Yann and Ballas, Nicolas},
  journal={arXiv preprint arXiv:2301.08243},
  year={2023}
}

@misc{girgis2025timeseriesjepapredictiveremote,
      title={Time-Series JEPA for Predictive Remote Control under Capacity-Limited Networks}, 
      author={Abanoub M. Girgis and Alvaro Valcarce and Mehdi Bennis},
      year={2025},
      eprint={2406.04853},
      archivePrefix={arXiv},
      primaryClass={cs.IT},
      url={https://arxiv.org/abs/2406.04853}, 
}

@article{Ennadir2025JointEG,
  title={Joint Embeddings Go Temporal},
  author={Sofiane Ennadir and Siavash Golkar and Leopoldo Sarra},
  journal={ArXiv},
  year={2025},
  volume={abs/2509.25449},
  url={https://api.semanticscholar.org/CorpusID:273754976}
}

@article{Wang2023CorrelatedTS,
  title={Correlated Time Series Self-Supervised Representation Learning via Spatiotemporal Bootstrapping},
  author={Lu-Xuan Wang and Lei Bai and Ziyue Li and Rui Zhao and Fugee Tsung},
  journal={2023 IEEE 19th International Conference on Automation Science and Engineering (CASE)},
  year={2023},
  pages={1-7},
  url={https://api.semanticscholar.org/CorpusID:259138815}
}

@inproceedings{Luo2026STOCKJEPAPL,
  title={STOCK-JEPA: Prior-Anchored Latent Revision Representation Learning in Equity Markets},
  author={Yi-Zhi Luo and Jia-He Yi and Jian-Hui Zhang and Shuo Sun},
  year={2026},
  url={https://api.semanticscholar.org/CorpusID:292662907}
}

@inproceedings{He2026SCJEPASL,
  title={SC-JEPA: Stabilizing Latent Predictive Learning for Time-Series Anomaly Prediction},
  author={Yanan He and Yunshi Wen and Xin Wang and Tengfei Ma},
  year={2026},
  url={https://api.semanticscholar.org/CorpusID:285286285}
}

@article{Han2023TheCA,
  title={The Capacity and Robustness Trade-Off: Revisiting the Channel Independent Strategy for Multivariate Time Series Forecasting},
  author={Lu Han and Han-Jia Ye and De-Chuan Zhan},
  journal={IEEE Transactions on Knowledge and Data Engineering},
  year={2023},
  volume={36},
  pages={7129-7142},
  url={https://api.semanticscholar.org/CorpusID:258060249}
}

@inproceedings{ICLR2026_0c6639f4,
 author = {HE, Cheng and Huang, Xu and Jiang, Gangwei and Li, Zhaoyi and Lian, Defu and Xie, Hong and Chen, Enhong and liang, xijie and Zhengzengrong and Lee, Patrick P. C.},
 booktitle = {International Conference on Learning Representations},
 editor = {C. Vondrick and B. Hariharan and C. Raffel and L. Pinto and D. Yang and A. Faust},
 pages = {7280--7309},
 title = {GTM: A General Time-series Model for Enhanced Representation Learning of Time-Series data},
 url = {https://proceedings.iclr.cc/paper_files/paper/2026/file/0c6639f49f01a8578675303ce0030233-Paper-Conference.pdf},
 volume = {2026},
 year = {2026}
}

@inproceedings{ICLR2026_ae3e1733,
 author = {Cheng, Hanyin and Wu, Xingjian and Shu, Yang and Rao, Zhongwen and Pan, Lujia and Yang, Bin and Guo, Guo},
 booktitle = {International Conference on Learning Representations},
 editor = {C. Vondrick and B. Hariharan and C. Raffel and L. Pinto and D. Yang and A. Faust},
 pages = {106657--106680},
 title = {CoRA: Boosting Time Series Foundation Models for Multivariate Forecasting through Correlation-aware Adapter},
 url = {https://proceedings.iclr.cc/paper_files/paper/2026/file/ae3e173398d6e43fea63cbcc16fbaa98-Paper-Conference.pdf},
 volume = {2026},
 year = {2026}
}

@inproceedings{NEURIPS2025_ecfb69ce,
 author = {He, Hui and Yi, Kun and Ma, Yuanchi and Zhang, Qi and Niu, Zhendong and Pang, Guansong},
 booktitle = {Advances in Neural Information Processing Systems},
 doi = {10.52202/085713-5405},
 editor = {D. Belgrave and C. Zhang and H. Lin and R. Pascanu and P. Koniusz and M. Ghassemi and N. Chen},
 pages = {161846--161879},
 publisher = {Curran Associates, Inc.},
 title = {SEMPO: Lightweight Foundation Models for Time Series Forecasting},
 url = {https://proceedings.neurips.cc/paper_files/paper/2025/file/ecfb69ce6be017deb5a926c2718f6bc1-Paper-Conference.pdf},
 volume = {38, Main Conference},
 year = {2025}
}

@InProceedings{pmlr-v267-liu25an,
  title = 	 {Moirai-{M}o{E}: Empowering Time Series Foundation Models with Sparse Mixture of Experts},
  author =       {Liu, Xu and Liu, Juncheng and Woo, Gerald and Aksu, Taha and Liang, Yuxuan and Zimmermann, Roger and Liu, Chenghao and Li, Junnan and Savarese, Silvio and Xiong, Caiming and Sahoo, Doyen},
  booktitle = 	 {Proceedings of the 42nd International Conference on Machine Learning},
  pages = 	 {38940--38962},
  year = 	 {2025},
  editor = 	 {Singh, Aarti and Fazel, Maryam and Hsu, Daniel and Lacoste-Julien, Simon and Berkenkamp, Felix and Maharaj, Tegan and Wagstaff, Kiri and Zhu, Jerry},
  volume = 	 {267},
  series = 	 {Proceedings of Machine Learning Research},
  month = 	 {13--19 Jul},
  publisher =    {PMLR},
  url = 	 {https://proceedings.mlr.press/v267/liu25an.html}
}

@misc{verdenius2024latpfnjointembeddingpredictive,
      title={LaT-PFN: A Joint Embedding Predictive Architecture for In-context Time-series Forecasting}, 
      author={Stijn Verdenius and Andrea Zerio and Roy L. M. Wang},
      year={2024},
      eprint={2405.10093},
      archivePrefix={arXiv},
      primaryClass={cs.LG},
      url={https://arxiv.org/abs/2405.10093}, 
}

@InProceedings{pmlr-v235-das24c,
  title = 	 {A decoder-only foundation model for time-series forecasting},
  author =       {Das, Abhimanyu and Kong, Weihao and Sen, Rajat and Zhou, Yichen},
  booktitle = 	 {Proceedings of the 41st International Conference on Machine Learning},
  pages = 	 {10148--10167},
  year = 	 {2024},
  editor = 	 {Salakhutdinov, Ruslan and Kolter, Zico and Heller, Katherine and Weller, Adrian and Oliver, Nuria and Scarlett, Jonathan and Berkenkamp, Felix},
  volume = 	 {235},
  series = 	 {Proceedings of Machine Learning Research},
  month = 	 {21--27 Jul},
  publisher =    {PMLR},
  url = 	 {https://proceedings.mlr.press/v235/das24c.html}
}

@misc{ansari2025chronos2univariateuniversalforecasting,
      title={Chronos-2: From Univariate to Universal Forecasting}, 
      author={Abdul Fatir Ansari and Oleksandr Shchur and Jaris Küken and Andreas Auer and Boran Han and Pedro Mercado and Syama Sundar Rangapuram and Huibin Shen and Lorenzo Stella and Xiyuan Zhang and Mononito Goswami and Shubham Kapoor and Danielle C. Maddix and Pablo Guerron and Tony Hu and Junming Yin and Nick Erickson and Prateek Mutalik Desai and Hao Wang and Huzefa Rangwala and George Karypis and Yuyang Wang and Michael Bohlke-Schneider},
      year={2025},
      eprint={2510.15821},
      archivePrefix={arXiv},
      primaryClass={cs.LG},
      url={https://arxiv.org/abs/2510.15821}, 
}

@InProceedings{pmlr-v235-goswami24a,
  title = 	 {{MOMENT}: A Family of Open Time-series Foundation Models},
  author =       {Goswami, Mononito and Szafer, Konrad and Choudhry, Arjun and Cai, Yifu and Li, Shuo and Dubrawski, Artur},
  booktitle = 	 {Proceedings of the 41st International Conference on Machine Learning},
  pages = 	 {16115--16152},
  year = 	 {2024},
  editor = 	 {Salakhutdinov, Ruslan and Kolter, Zico and Heller, Katherine and Weller, Adrian and Oliver, Nuria and Scarlett, Jonathan and Berkenkamp, Felix},
  volume = 	 {235},
  series = 	 {Proceedings of Machine Learning Research},
  month = 	 {21--27 Jul},
  publisher =    {PMLR},
  url = 	 {https://proceedings.mlr.press/v235/goswami24a.html}
}

@misc{fu2025frequencymaskedembeddinginferencenoncontrastive,
      title={Frequency-Masked Embedding Inference: A Non-Contrastive Approach for Time Series Representation Learning}, 
      author={En Fu and Yanyan Hu},
      year={2025},
      eprint={2412.20790},
      archivePrefix={arXiv},
      primaryClass={cs.LG},
      url={https://arxiv.org/abs/2412.20790}, 
}

@misc{shi2025kronosfoundationmodellanguage,
      title={Kronos: A Foundation Model for the Language of Financial Markets}, 
      author={Yu Shi and Zongliang Fu and Shuo Chen and Bohan Zhao and Wei Xu and Changshui Zhang and Jian Li},
      year={2025},
      eprint={2508.02739},
      archivePrefix={arXiv},
      primaryClass={q-fin.ST},
      url={https://arxiv.org/abs/2508.02739}, 
}

@article{Rahimikia2025ReVisitingTS,
  title={Re(Visiting) Time Series Foundation Models in Finance},
  author={Eghbal Rahimikia and Hao Ni and Weiguan Wang},
  journal={ArXiv},
  year={2025},
  volume={abs/2511.18578},
  url={https://api.semanticscholar.org/CorpusID:283243969}
}

@misc{alonso2026pretrainedtimeseriesfoundationmodels,
      title={Pretrained Time-Series Foundation Models for Financial Return Forecasting}, 
      author={Miquel Noguer I Alonso and Rodolfo Pereira Franklin},
      year={2026},
      eprint={2606.27100},
      archivePrefix={arXiv},
      primaryClass={q-fin.MF},
      url={https://arxiv.org/abs/2606.27100}, 
}

@misc{rasul2024lagllamafoundationmodelsprobabilistic,
      title={Lag-Llama: Towards Foundation Models for Probabilistic Time Series Forecasting}, 
      author={Kashif Rasul and Arjun Ashok and Andrew Robert Williams and Hena Ghonia and Rishika Bhagwatkar and Arian Khorasani and Mohammad Javad Darvishi Bayazi and George Adamopoulos and Roland Riachi and Nadhir Hassen and Marin Biloš and Sahil Garg and Anderson Schneider and Nicolas Chapados and Alexandre Drouin and Valentina Zantedeschi and Yuriy Nevmyvaka and Irina Rish},
      year={2024},
      eprint={2310.08278},
      archivePrefix={arXiv},
      primaryClass={cs.LG},
      url={https://arxiv.org/abs/2310.08278}, 
}

@misc{lin2024nutimenumericallymultiscaledembedding,
      title={NuTime: Numerically Multi-Scaled Embedding for Large-Scale Time-Series Pretraining}, 
      author={Chenguo Lin and Xumeng Wen and Wei Cao and Congrui Huang and Jiang Bian and Stephen Lin and Zhirong Wu},
      year={2024},
      eprint={2310.07402},
      archivePrefix={arXiv},
      primaryClass={cs.LG},
      url={https://arxiv.org/abs/2310.07402}, 
}

@article{Yang2020QlibAA,
  title={Qlib: An AI-oriented Quantitative Investment Platform},
  author={Xiao Yang and Wei-Qing Liu and Dong Zhou and Jiang Bian and Tie-Yan Liu},
  journal={ArXiv},
  year={2020},
  volume={abs/2009.11189},
  url={https://api.semanticscholar.org/CorpusID:221856639}
}

@misc{oord2019representationlearningcontrastivepredictive,
      title={Representation Learning with Contrastive Predictive Coding}, 
      author={Aaron van den Oord and Yazhe Li and Oriol Vinyals},
      year={2019},
      eprint={1807.03748},
      archivePrefix={arXiv},
      primaryClass={cs.LG},
      url={https://arxiv.org/abs/1807.03748}, 
}

@inproceedings{woo2022iclr-cost,
  title     = {{CoST: Contrastive Learning of Disentangled Seasonal-Trend Representations for Time Series Forecasting}},
  author    = {Woo, Gerald and Liu, Chenghao and Sahoo, Doyen and Kumar, Akshat and Hoi, Steven},
  booktitle = {International Conference on Learning Representations},
  year      = {2022},
  url       = {https://mlanthology.org/iclr/2022/woo2022iclr-cost/}
}

@inproceedings{tonekaboni2021iclr-unsupervised,
  title     = {{Unsupervised Representation Learning for Time Series with Temporal Neighborhood Coding}},
  author    = {Tonekaboni, Sana and Eytan, Danny and Goldenberg, Anna},
  booktitle = {International Conference on Learning Representations},
  year      = {2021},
  url       = {https://mlanthology.org/iclr/2021/tonekaboni2021iclr-unsupervised/}
}

@inproceedings{liu2024timesurl,
  title={Timesurl: Self-supervised contrastive learning for universal time series representation learning},
  author={Liu, Jiexi and Chen, Songcan},
  booktitle={Proceedings of the AAAI Conference on Artificial Intelligence},
  volume={38},
  number={12},
  pages={13918--13926},
  year={2024}
}

@inproceedings{Yuqietal-2023-PatchTST,
  title     = {A Time Series is Worth 64 Words: Long-term Forecasting with Transformers},
  author    = {Nie, Yuqi and
               H. Nguyen, Nam and
               Sinthong, Phanwadee and 
               Kalagnanam, Jayant},
  booktitle = {International Conference on Learning Representations},
  year      = {2023}
}

@inproceedings{dong2023simmtm,
  title={SimMTM: A Simple Pre-Training Framework for Masked Time-Series Modeling},
  author={Jiaxiang Dong and Haixu Wu and Haoran Zhang and Li Zhang and Jianmin Wang and Mingsheng Long},
  booktitle={Advances in Neural Information Processing Systems},
  year={2023}
}

@inproceedings{merchant2026towards,
  title={Towards Financial World Modeling},
  author={Merchant, Humzah and Guthrie, Alec and Mahns, Simon and Balestriero, Randall and Levy, Bradford},
  booktitle={The Fortieth Annual Conference on Neural Information Processing Systems},
  year={2026},
  url={https://arxiv.org/abs/2610.09048}
}

@article{xu2021hist,
  title={HIST: A Graph-based Framework for Stock Trend Forecasting via Mining Concept-Oriented Shared Information},
  author={Xu, Wentao and Liu, Weiqing and Wang, Lewen and Xia, Yingce and Bian, Jiang and Yin, Jian and Liu, Tie-Yan},
  journal={arXiv preprint arXiv:2110.13716},
  year={2021}
}

@inproceedings{li2024master,
  title={Master: Market-guided stock transformer for stock price forecasting},
  author={Li, Tong and Liu, Zhaoyang and Shen, Yanyan and Wang, Xue and Chen, Haokun and Huang, Sen},
  booktitle={Proceedings of the AAAI Conference on Artificial Intelligence},
  volume={38},
  number={1},
  pages={162--170},
  year={2024}
}

@article{sharpe1964capital,
  title={Capital asset prices: A theory of market equilibrium under conditions of risk},
  author={Sharpe, William F},
  journal={The journal of finance},
  volume={19},
  number={3},
  pages={425--442},
  year={1964},
  publisher={Wiley Online Library}
}

@article{fama1993common,
  title={Common risk factors in the returns on stocks and bonds},
  author={Fama, Eugene F and French, Kenneth R},
  journal={Journal of financial economics},
  volume={33},
  number={1},
  pages={3--56},
  year={1993},
  publisher={Elsevier}
}

@article{hasbrouck2001common,
  title={Common factors in prices, order flows, and liquidity},
  author={Hasbrouck, Joel and Seppi, Duane J},
  journal={Journal of financial Economics},
  volume={59},
  number={3},
  pages={383--411},
  year={2001},
  publisher={Elsevier}
}

@article{gu2021autoencoder,
  title={Autoencoder asset pricing models},
  author={Gu, Shihao and Kelly, Bryan and Xiu, Dacheng},
  journal={Journal of Econometrics},
  volume={222},
  number={1},
  pages={429--450},
  year={2021},
  publisher={Elsevier}
}

@inproceedings{duan2022factorvae,
  title={Factorvae: A probabilistic dynamic factor model based on variational autoencoder for predicting cross-sectional stock returns},
  author={Duan, Yitong and Wang, Lei and Zhang, Qizhong and Li, Jian},
  booktitle={Proceedings of the AAAI conference on artificial intelligence},
  volume={36},
  number={4},
  pages={4468--4476},
  year={2022}
}

@article{vaswani2017attention,
  title={Attention is all you need},
  author={Vaswani, Ashish and Shazeer, Noam and Parmar, Niki and Uszkoreit, Jakob and Jones, Llion and Gomez, Aidan N and Kaiser, {\L}ukasz and Polosukhin, Illia},
  journal={Advances in neural information processing systems},
  volume={30},
  year={2017}
}

@inproceedings{yue2022ts2vec,
  title={Ts2vec: Towards universal representation of time series},
  author={Yue, Zhihan and Wang, Yujing and Duan, Juanyong and Yang, Tianmeng and Huang, Congrui and Tong, Yunhai and Xu, Bixiong},
  booktitle={Proceedings of the AAAI conference on artificial intelligence},
  volume={36},
  number={8},
  pages={8980--8987},
  year={2022}
}
